%% file: manuscript.tex
\documentclass[10pt,journal,compsoc]{IEEEtran}
\ifCLASSOPTIONcompsoc
\usepackage[nocompress]{cite}
\else
\usepackage{cite}
\fi
\ifCLASSINFOpdf
\else
\fi

\usepackage{epsfig}
\usepackage{graphicx}
\usepackage{amsmath}
\usepackage{amssymb}
\usepackage{enumitem}

\usepackage[pagebackref=true,breaklinks=true,letterpaper=true,colorlinks,bookmarks=false,urlcolor=red,citecolor=blue]{hyperref}
\usepackage{gensymb,graphics,subfigure,inputenc}
\usepackage[linesnumbered,algo2e,boxed]{algorithm2e}
\usepackage{ragged2e} 
\graphicspath{{Figure/}}
\usepackage[figuresright]{rotating}
\usepackage{amsmath,amssymb,textcomp,subfigure,multirow,upgreek}
\usepackage{booktabs}
\usepackage{color,mdwlist}

\usepackage{makecell}
\usepackage{graphicx}
\usepackage{utfsym}
\usepackage{arydshln}
\usepackage{cleveref}
\usepackage{threeparttable,booktabs}
\usepackage{subcaption}
\usepackage{longtable}

\usepackage{wasysym}
\usepackage{utfsym}
\usepackage{amsbsy}
\usepackage{pifont}
\newcommand{\cmark}{\ding{51}}
\newcommand{\xmark}{\ding{55}}

\usepackage{threeparttablex} 
\usepackage{array}           

\definecolor{ManuscriptRed}{HTML}{8B0000}     
\definecolor{ManuscriptBlack}{HTML}{000000}     
\newcommand{\change}[1]{{\color{ManuscriptBlack}#1}}
\newcommand{\changeminor}[1]{{\color{ManuscriptBlack}#1}}

\usepackage{graphicx,fontawesome}
\graphicspath{{Figures/}}

\usepackage{caption}
\usepackage{enumitem}
\setitemize{noitemsep,topsep=0pt,parsep=0pt,partopsep=0pt}


\begin{document}
	%
    \title{Face De-Identification: A Domain-Centric Survey from Capture to Processing}
	\author{Hui~Wei, Hao~Yu, and Guoying~Zhao, Fellow, IEEE
		\IEEEcompsocitemizethanks{
            \IEEEcompsocthanksitem Hui Wei, Hao~Yu, and Guoying Zhao are with the ELLIS Institute Finland, Espoo, 02150, Finland, and also with the Center for Machine Vision and Signal Analysis, University of Oulu, 90570 Oulu, Finland. \protect
			\IEEEcompsocthanksitem Guoying~Zhao is the corresponding author. E-mail: guoying.zhao@oulu.fi \protect
		}
	}
	
	%
	%
	
	\markboth{submit to IEEE Transactions on Pattern Analysis and Machine Intelligence}%
	{Shell \MakeLowercase{\textit{et al.}}: Bare Demo of IEEEtran.cls for Computer Society Journals}
	%



	\IEEEtitleabstractindextext{%
		\justify

\begin{abstract}
Face de-identification (De-ID) aims to remove or conceal personally identifiable facial features in images or videos to prevent identity recognition while preserving utility for downstream tasks. 
With the rising emphasis on data privacy and responsible AI, face De-ID has emerged as an active research area spanning computer vision and privacy-preserving communities.
Early approaches, and many contemporary ones, operate in the \textbf{digital} domain by modifying pixel-level or appearance-level features through post-capture processing. 
Recent advances extend face De-ID beyond post-processing by integrating privacy mechanisms directly into \textbf{sensors} during image acquisition, bridging sensing systems and downstream vision algorithms. 
In parallel, \textbf{physical}-domain methods explore wearable accessories and materials that conceal identity information in real-world environments prior to capture.
In this survey, we present the first unified \change{overview} that spans the full data acquisition pipeline, encompassing the physical, sensor, and digital domains. 
Through this domain-centric lens, we systematically analyze current methodologies, technical progress, and the distinct challenges inherent to each stage.
We then review and organize existing evaluation protocols, examining current practices and highlighting the critical need for standardized, comprehensive benchmarks. 
Finally, we identify key open problems and outline emerging research directions to guide future work in this rapidly evolving field. 
To support ongoing research, we maintain a project page that organizes relevant literature with collected datasets and open source code: \url{https://github.com/CV-AC/Awesome-FaceDe-ID}.

\end{abstract}
		
		\begin{IEEEkeywords}
			Face De-identification, Utility Preservation, Evaluation Protocol, Privacy Protection, Literature Survey.
	\end{IEEEkeywords}}

	\maketitle

	\IEEEdisplaynontitleabstractindextext

	%
	\IEEEpeerreviewmaketitle


	%
	%
	%
	%

\IEEEraisesectionheading{\section{Introduction}
	\label{sec:introduction}}

\IEEEPARstart{H}{uman} face contains abundant identity-related information, making it a cornerstone of biometric recognition systems in modern society~\cite{zhao2003face,guo20233d}. However, the widespread collection and utilization of facial data pose significant privacy, security, and ethical concerns~\cite{mi2024privacy,laishram2025toward}. 
Recent regulatory frameworks, most notably the EU AI Act (Regulation (EU) 2024/1689), underscore that the misuse of facial data constitutes a severe infringement of fundamental rights, including the right to privacy~\cite{EUAIAct2024}.
\change{To mitigate these risks, the research community has developed \textbf{face de-identification (De-ID)}, which is defined as the class of image- or video-domain transformations that conceal personally identifiable facial information such that the depicted subject cannot be recognized, by either an automated face recognizer or a human observer, while non-identity attributes required by downstream applications (\emph{e.g.}, medical analysis~\cite{yang2022digital} and attribute recognition~\cite{cai2024disguise}) remain usable. }

\begin{figure}
    \centering
    \includegraphics[width=1.0\linewidth]{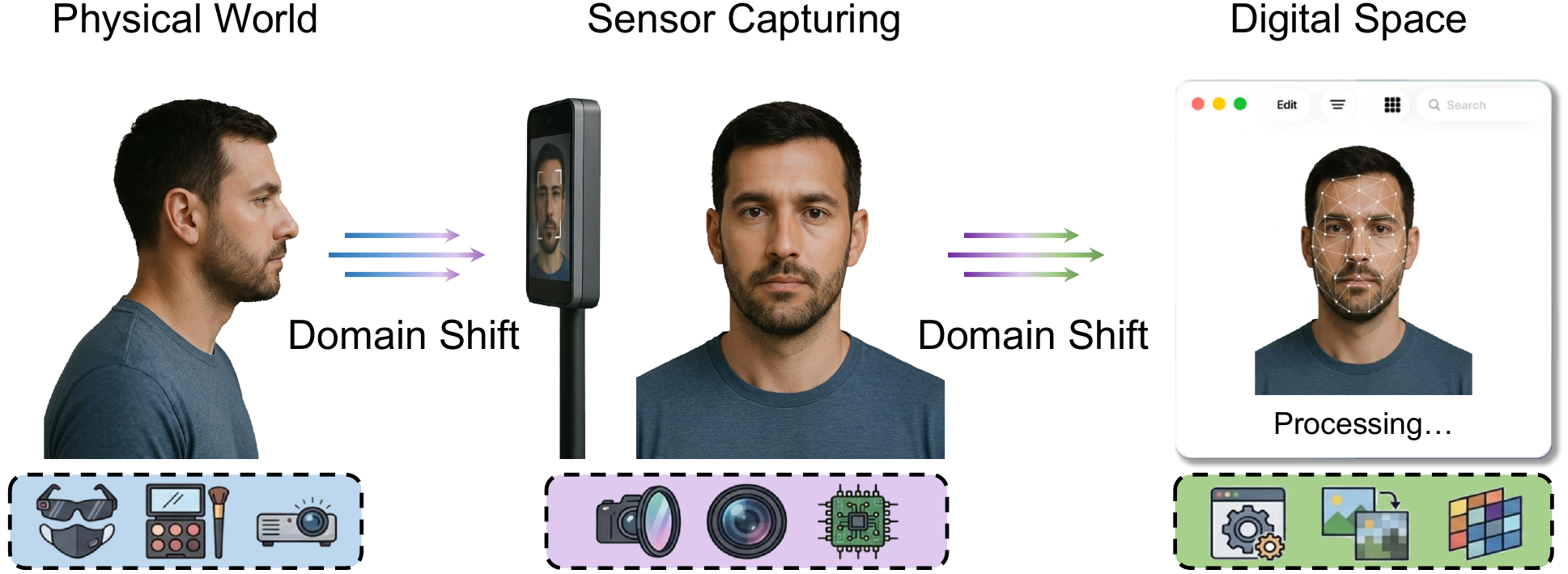}
    \caption{\textbf{Overall perspective of this face De-ID survey paper}. The process begins in \textbf{physical} world, where identity cues are modified directly on the subject through adversarial wearables, cosmetics, or projection-based interventions. It then transitions through the \textbf{sensor}-capturing stage, where privacy-preserving optics and computational imaging encode De-ID at the sensor level. Finally, in \textbf{digital} space, post-capture algorithms apply pixel- or feature-level transformations to conceal identity while retaining utility.}
    \label{fig:perspective}
    \vspace{-5mm}
\end{figure}

In recent years, face De-ID has been widely studied and has expanded from purely digital treatments to encompass sensor-integrated and physical-world interventions (see Fig.~\ref{fig:perspective}). 
Early face De-ID efforts predominantly operated in the \textbf{digital domain}, manipulating recorded imagery via hand-crafted filters (e.g., blurring, pixelation)~\cite{hudson1996techniques} or statistical anonymization mechanisms such as \textit{k}-Same family~\cite{newton2005preserving}. 
Subsequent advances leveraged adversarial perturbations~\cite{NEURIPS2022_dccbeb7a,salar2025enhancing} and generative models~\cite{zhai2022a3gan,kung2025face} to synthesize photorealistic yet privacy-safe faces.
While these digital approaches achieved remarkable fidelity and controllability, they remain post hoc interventions, exposing raw sensitive data prior to De-ID~\cite{pittaluga2015privacy}.
To mitigate this vulnerability, recent research has shifted privacy protection {into the imaging process}. \textbf{Sensor domain} face De-ID paradigms integrate optical and computational imaging designs that suppress identity cues at capture time, embedding privacy filters in optics or sensors to eliminate identifiable cues before digital conversion (\emph{e.g.}, PrivacyOptics~\cite{lopez2024privacy}). In parallel, \textbf{physical domain} strategies extend face De-ID into the real world by modifying subjects’ appearance through adversarial wearables~\cite{sharif2016accessorize}, cosmetics~\cite{yin2021adv}, or projected light patterns~\cite{liu2025projattacker} that confuse recognition systems under unconstrained environments. Collectively, these directions represent a continuum from post-capture face De-ID to sensor-level and physical-level protection, each offering distinct advantages and challenges regarding effectiveness, utility, and realism.

\begin{figure}[t]
    \centering
    \includegraphics[width=1.0\linewidth]{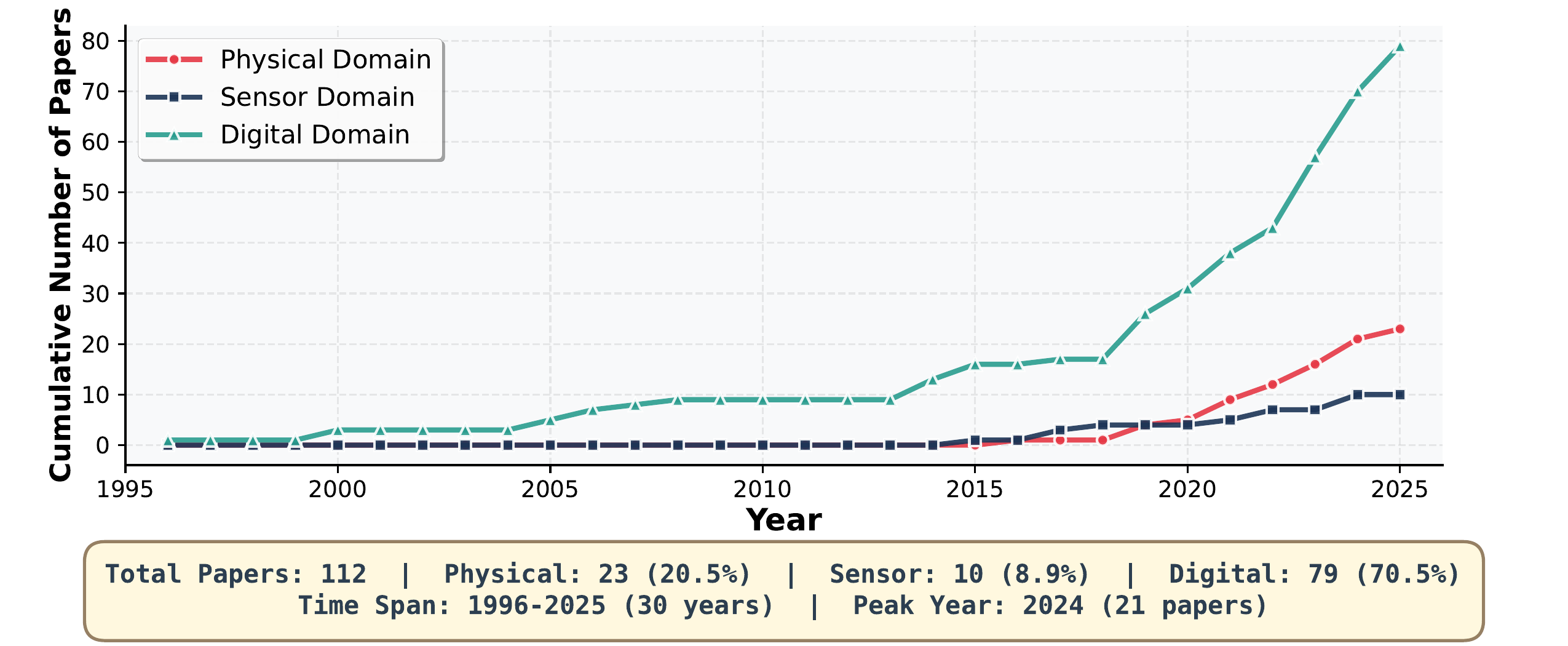}
    \caption{\textbf{Cumulative growth of face De-ID methods across domains}. Data collected via Google Scholar using keywords: "face de-identification", "face privacy preserving", and "face adversarial attack".}
    \label{fig:evolution}
    \vspace{-5mm}
\end{figure}

\begin{figure*}[t]
\centering    
\subfigure[Relationship between face De-ID and other privacy-preserving paradigms.]{\label{figure:taxonomy}\includegraphics[height=29.0mm]{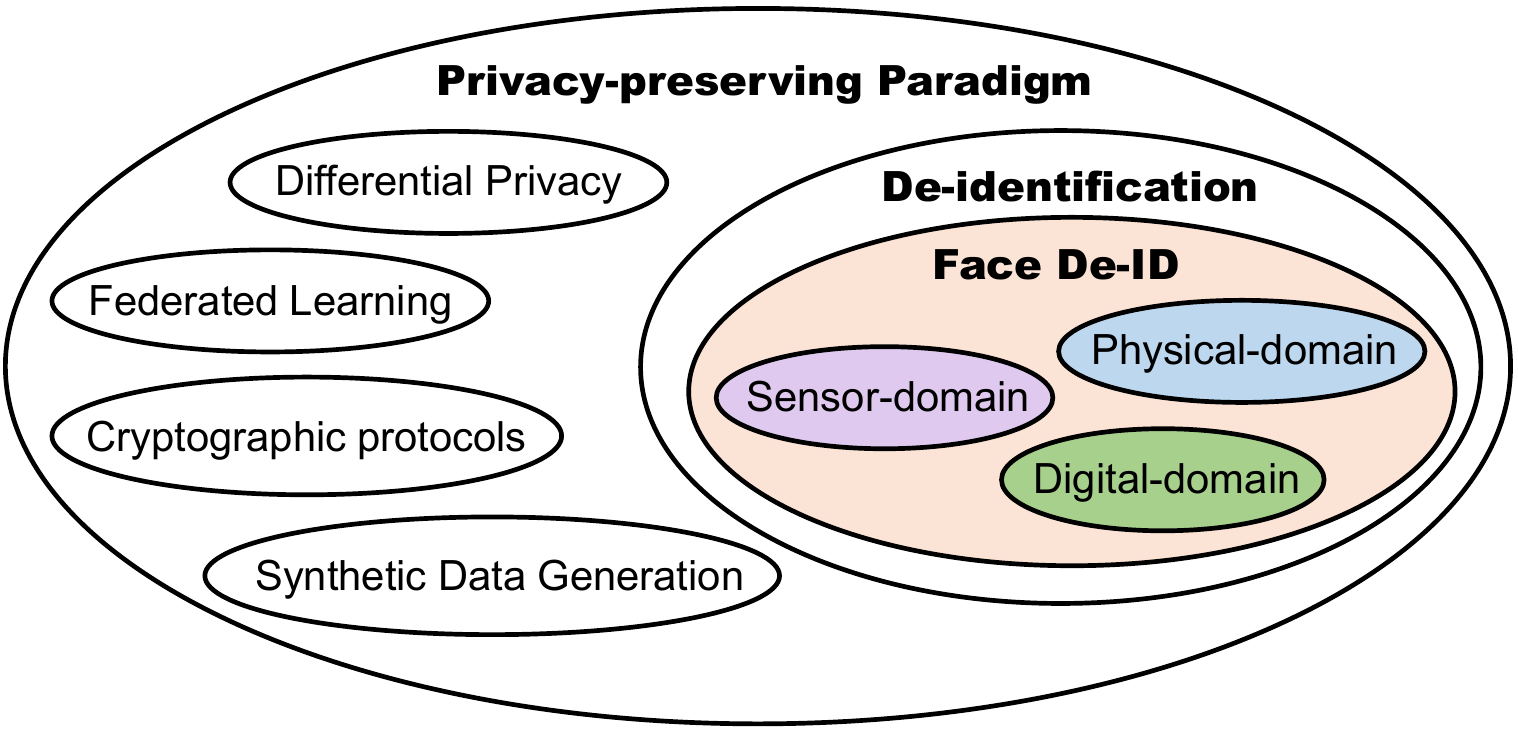}} 
\hspace{1mm}      
\rule{0.5pt}{30mm} 
\hspace{1mm}  
\subfigure[Chronological overview of representative face De-ID works. The timeline highlights the shift from early digital methods to sensor-based and physical solutions.]{\label{fig:timeline}\includegraphics[height=30.8mm]{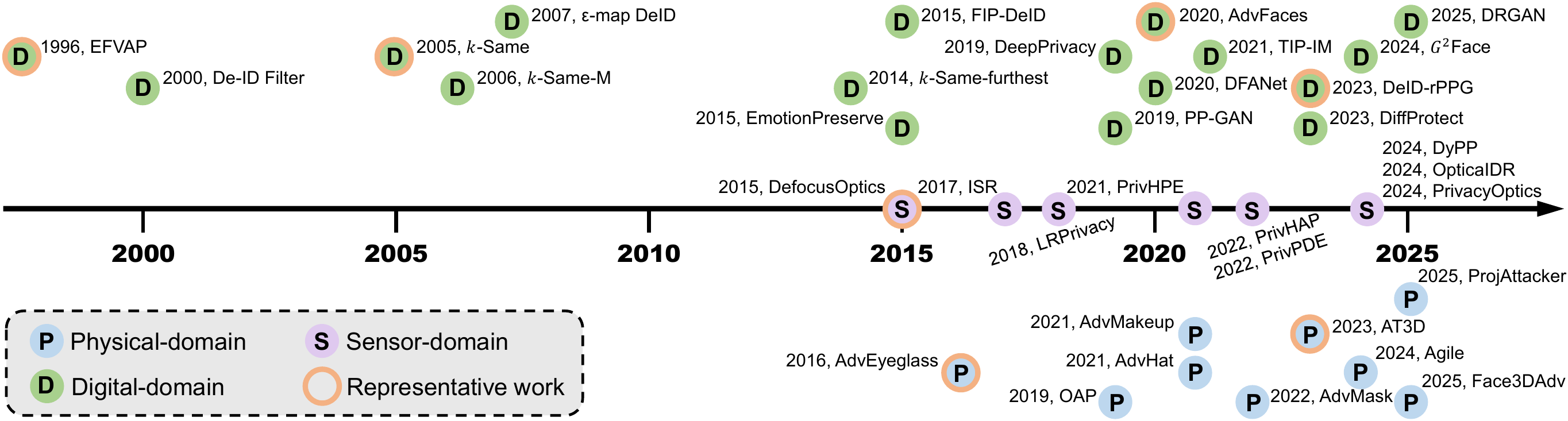}}
\vspace{-1mm}
\caption{\textbf{Conceptual} (left) and \textbf{historical} (right) overview of face De-ID, clarifying the paper’s domain-centric taxonomy.}
\label{figure:preliminary}
  \vspace{-0.4cm}
\end{figure*}

This paper provides a unique and up-to-date survey of the rapidly growing area of face De-ID (Fig.~\ref{fig:evolution} illustrates the growth trajectory and domain-specific research trends). Although several survey papers have reviewed aspects of face De-ID, existing works remain fragmented and narrow in scope (see TABLE~\ref{tab:survey}). Specifically, most prior surveys concentrate only on digital domain techniques~\cite{liu2022privacy,cheng2022benchmarking}, overlooking the rapid evolution of sensor-based and physical face De-ID methods that increasingly shape real-world privacy protection. Moreover, current surveys rarely provide evaluation protocol analysis~\cite{hasan2023presentation,cao2024face}, resulting in insufficient overview of the datasets and metrics that are critical for advancing this field.

To address these gaps, this survey presents a comprehensive review and critical analysis of face De-ID techniques from \textbf{physical}, \textbf{sensor}, and \textbf{digital} perspectives, spanning more than three decades of research. Our goal is to bridge the fragmented literature across these domains, establish a unified taxonomy, and highlight both the fragmentation of evaluation protocols in this field and the emerging convergence between optical design, adversarial learning, and generative synthesis for privacy preservation. Specifically, this paper makes the following contributions:

\begin{itemize}
    \item We provide the first \textbf{domain-centric taxonomy} of face De-ID methods, systematically categorizing approaches according to where privacy transformation occurs: before, during, or after image capture, and summarizing representative methodologies within each domain.
    \item We perform an \textbf{in-depth analysis of methodologies and trends} across physical, sensor-based, and digital face De-ID, highlighting their design principles, advantages, and limitations.
    \item We review and summarize \textbf{evaluation protocols}, emphasizing the need for unified assessment frameworks that jointly measure privacy protection, utility preservation, and visual quality.
    \item We identify \textbf{open challenges and future directions}, including cross-modal privacy preservation, foundation model level face De-ID, verifiable and reversible De-ID, and fairness-aware privacy guarantees.
\end{itemize}

The remainder of this paper is organized as follows. Sec.~\ref{sec:preliminaries} establishes conceptual foundations. Sec.~\ref{sec:method} reviews methodologies across the three domains. Sec.~\ref{sec:protocol} systematizes evaluation protocols. Sec.~\ref{sec:discussion} outlines open problems and Sec.~\ref{sec:conclusion} concludes the survey.

\begin{table}[t]
\caption{\textbf{Summary of existing survey papers related to face De-ID}. Prior works predominantly focus on digital-domain methods, with limited exploration of sensor-based or physical de-identification. Our survey provides the first unified, domain-centric overview that spans all three levels and includes evaluation protocol analysis.}
\label{tab:survey}
\setlength{\tabcolsep}{3pt}
\resizebox{1.0\linewidth}{!}{
\begin{tabular}{ccccccl}
\toprule
Paper & Year & Physical & Sensor & Digital &   Protocol  & Perspective\\
\midrule
\cite{padilla2015visual} & 2015 & \xmark & \cmark & \cmark & \xmark & Camera \\
\cite{ribaric2015overview} & 2015 & \xmark & \cmark & \cmark & \xmark & Image/video \\
\cite{ribaric2016identification} &  2016 &  \xmark & \cmark & \cmark & \xmark & Multimedia \\
\cite{meden2021privacy} &  \change{2021} & \change{\xmark} & \change{\cmark} & \change{\cmark} & \change{\xmark} & \change{Pipeline-level}\\
\cite{liu2022privacy} & 2022 & \xmark &  \xmark &  \cmark & \xmark &  Social network\\
\cite{cheng2022benchmarking} & 2022 & \xmark &  \xmark &  \cmark & \xmark & 3D face De-ID\\
\cite{hasan2023presentation} & 2023 & \cmark & \xmark & \xmark & \xmark & Presentation-level \\
\cite{cao2024face} & 2024 & \xmark & \xmark & \cmark & \xmark & Processing level\\
\cite{khan2024person} & 2024 & \xmark & \xmark & \cmark & \xmark & Person De-ID \\
\cite{wen2024face} & 2024 & \xmark & \xmark & \cmark & \xmark & Digital era\\ 
\cite{goswami2025faces} & 2025 & \xmark & \xmark & \cmark &   \xmark & Generative \\
\cite{park2025privacy} & 2025 & \xmark & \xmark & \cmark & \xmark & Deepfake  \\
\cite{laishram2025toward} & 2025 &  \xmark & \cmark & \cmark &  \xmark & Leakage/solution\\
Ours & --- & \cmark  & \cmark  & \cmark  & \cmark & Domain-centric \\
\bottomrule
\end{tabular}
}
\vspace{-4mm}
\end{table}

\section{Preliminaries}
	\label{sec:preliminaries}
Research on face De-ID traces back to the early era of computational face analysis. The first successful demonstration of automatic face recognition, Eigenfaces~\cite{turk1991eigenfaces} by Turk and Pentland in 1991, highlighted both the feasibility of machine-based identification and the accompanying privacy risks.
Shortly thereafter, the first dedicated De-ID technique, EFVAP~\cite{hudson1996techniques}, was introduced in 1996, marking the beginning of systematic efforts to suppress identifiable facial information. 
\change{Since then, the field has expanded into a broad research direction. Fig.~\ref{fig:timeline} presents a curated chronological trajectory of representative face De-ID methods, revealing a clear shift in research emphasis over time. A work appears in the timeline if it satisfies at least two of the following criteria: \emph{(i)} it introduced a methodological primitive subsequently adopted by other works (\emph{e.g.}, $k$-Same~\cite{newton2005preserving}, AdvEyeglass~\cite{sharif2016accessorize}, DefocusOptics~\cite{pittaluga2015privacy}); \emph{(ii)} it is the first reported method in its sub-category (\emph{e.g.}, EFVAP~\cite{hudson1996techniques}); \emph{(iii)} it represents the current state of the art or a distinct subsequent generation within a sub-category (\emph{e.g.}, AT3D~\cite{yang2023towards}, $G^2$Face~\cite{yang2024g}); or \emph{(iv)} it has high scholarly impact (citations $\geq 200$ for works published before 2022 and $\geq 50$ from 2022 onwards). Exhaustive coverage of all 112 surveyed methods, including those not selected for the timeline, is provided in TABLE~\ref{tab:methods_summary}.}

\vspace{-3mm}
\subsection{Face De-ID \textit{vs}. Privacy-Preserving}

As shown in Fig.~\ref{figure:taxonomy}, face De-ID represents one branch within a broader family of privacy-preserving methodologies. 
More broadly, privacy-preserving methodologies comprise complementary paradigms, including differential privacy~\cite{zhao2022survey}, federated learning~\cite{huang2024federated}, cryptographic protocols~\cite{basin2022tamarin}, synthetic data generation~\cite{jia2025balancing}, and de-identification~\cite{wen2024face,hanisch2025anonymization}, each of which addresses distinct dimensions of data protection. 
\change{Face De-ID is a vision-centric technique that complements but does not replace other privacy paradigms.
Recognition in video and surveillance further exploits complementary modalities including body shape, gait, and other behavioral biometrics (\emph{e.g.}, voice, eye gaze, hand motions). Face De-ID is thus one component of a broader \emph{person} De-ID pipeline.}

\vspace{-3mm}
\subsection{Face De-ID \textit{vs}. Face Identification}

Face identification maximizes the discriminability of identity-bearing cues, whereas face De-ID suppresses those cues while preserving non-identity information. Most face De-ID approaches therefore assume a strong face recognition (FR) adversary as their primary threat model. Modern FR systems extract discriminative embeddings optimized for intra-class compactness and inter-class separability through metric learning objectives. Representative architectures include FaceNet~\cite{schroff2015facenet}, which pioneered triplet-loss deep metric learning; SphereFace~\cite{liu2017sphereface}, which introduced angular softmax for hypersphere embedding; ArcFace~\cite{deng2019arcface}, the most prevalent in De-ID research owing to its additive angular margin loss; and AdaFace~\cite{kim2022adaface}, which adapts margins to image quality. These systems typically operate on faces localized by detectors such as RetinaFace~\cite{deng2020retinaface}.

This adversarial relationship runs throughout the De-ID pipeline: physical methods optimize perturbations to fool FR models in the real world~\cite{yang2023towards}; sensor-based approaches design optics that suppress identity-discriminative frequencies~\cite{pan2024opticaldr}; and digital methods incorporate pre-trained FR models as verifiers in their training objectives~\cite{maximov2020ciagan}. With FR systems defining both the attack surface and the evaluation criteria, face identification is foundational to designing and benchmarking De-ID techniques.

\vspace{-3mm}
\subsection{Problem Formulation}
\label{sec:problem_formulation}

\change{
Let $\mathbf{I} \in \mathbb{R}^{H \times W \times 3}$ denote an input face image (or video frame) containing personally identifiable information about an individual. A face De-ID method seeks a transformation $\mathcal{F}(\cdot)$ that produces a de-identified output $\tilde{\mathbf{I}} = \mathcal{F}(\mathbf{I})$ in which identity-related cues are concealed while non-identity, task-relevant information is preserved. For applications that additionally require authorized recovery (\emph{e.g.}, forensic investigation, medical audit, legal disclosure), $\mathcal{F}$ is paired with a key-conditioned inverse $\mathcal{R}(\cdot\,;\,k)$ such that $\mathcal{R}(\tilde{\mathbf{I}};\,k) \approx \mathbf{I}$ only when $k$ is the legitimate recovery key.

Formally, we cast face De-ID as the joint \emph{maximization} of privacy, utility, visual-quality, and reversibility \emph{scores}:
\begin{equation}
\max_{\mathcal{F},\,\mathcal{R}} \;
\lambda_{p}\mathcal{P}
+ \lambda_{u}\mathcal{U}
+ \lambda_{q}\mathcal{Q}
+ \delta_{\text{rev}}\,\lambda_{r}\mathcal{S},
\label{eq:objective}
\end{equation}
where $\mathcal{P}(\mathbf{I},\tilde{\mathbf{I}})$, $\mathcal{U}(\mathbf{I},\tilde{\mathbf{I}})$, $\mathcal{Q}(\tilde{\mathbf{I}})$, and $\mathcal{S}(\mathbf{I},\tilde{\mathbf{I}},\mathcal{R};\,k)$ denote privacy, utility, visual-quality, and key-conditioned reversibility \emph{scores} (each to be maximized), $\lambda_{p},\lambda_{u},\lambda_{q},\lambda_{r} \geq 0$ are trade-off weights, and $\delta_{\text{rev}} \in \{0,1\}$ is an application-level switch that enables the reversibility term when authorized recovery is required and disables it otherwise. The expected solution should satisfy the following criteria:

\begin{enumerate}
    \item \textbf{Privacy Protection:} The de-identified $\tilde{\mathbf{I}}$ should not match any other identity. Formally, for any recognizer $f_{\text{id}}$,
    $\Pr\!\left(f_{\text{id}}(\tilde{\mathbf{I}}) = f_{\text{id}}(\mathbf{I}_i)\right) \approx 0,\ \forall i$,
    where $\mathbf{I}_i$ denotes any image of identity $i$. This prevents both re-identification to the original and misattribution to others.

    \item \textbf{Utility Preservation:} For a set of downstream tasks $\mathcal{T} = \{t_1,\ldots,t_n\}$, the task-specific predictors $f_{t_j}$ should maintain consistent outputs before and after face De-ID, \emph{i.e.}, $f_{t_j}(\tilde{\mathbf{I}}) \approx f_{t_j}(\mathbf{I})$.

    \item \textbf{Visual Fidelity and Naturalness:} The de-identified image should remain photorealistic and free from perceptible artifacts, ensuring usability and social acceptability in real applications.

    \item \textbf{Reversibility:} When $\delta_{\text{rev}} = 1$, there should exist a key-conditioned inverse $\mathcal{R}(\cdot\,;\,k)$ satisfying $\mathcal{R}(\tilde{\mathbf{I}};\,k) \approx \mathbf{I}$ for the legitimate key while $\Pr\!\left(\mathcal{R}(\tilde{\mathbf{I}};\,k') \approx \mathbf{I}\right) \to 0$ for any $k' \neq k$. Reversibility is orthogonal to privacy: a method may be strongly de-identifying yet reversible (\emph{e.g.}, password-conditioned generative inversion), strongly de-identifying and irreversible, or neither. 
\end{enumerate}
}

\vspace{-3mm}
\subsection{Taxonomy: A Domain-Centric Perspective}
\label{subsec:Taxonomy}
Face De-ID methods are fundamentally characterized by \emph{where} in the imaging pipeline the privacy transformation occurs. This survey adopts a domain-centric taxonomy with three classes, \textbf{physical}, \textbf{sensor}, and \textbf{digital}, each embodying distinct design constraints and trade-offs.

\noindent\textbf{Physical Domain.}
A transformation $\mathcal{F}_{p}$ alters the subject's real-world appearance before capture (\emph{e.g.}, adversarial accessories, makeup, light projection, 3D mask) so that the captured image $\tilde{\mathbf{I}}$ inherently exhibits suppressed identity. These methods strongly protect against raw-data exposure but offer limited controllability and may be conspicuous.

\noindent\textbf{Sensor Domain.}
Imaging components $\mathcal{F}_{s}$ embed privacy during acquisition through optical coding, sensor-level transformations, or computational imaging (\emph{e.g.}, learned phase masks, PSF manipulation), encoding $\tilde{\mathbf{I}} = \mathcal{F}_{s}(\mathbf{I})$ in the raw measurement space so that no identifiable face is produced. They provide strong trust-boundary guarantees but face stringent hardware constraints (latency, resolution, power) and deployment challenges.

\noindent\textbf{Digital Domain.}
Post-capture methods operate on recorded imagery, modifying identity through filtering, mapping, adversarial perturbations, or generative synthesis (\emph{e.g.}, GANs~\cite{goodfellow2014generative}, diffusion models~\cite{ho2020denoising}) to obtain $\tilde{\mathbf{I}} = \mathcal{F}_{d}(\mathbf{I})$. They are the most flexible and widely adopted, enabling fine-grained control and high fidelity, but expose raw data before De-ID and remain most vulnerable to re-identification.

\vspace{-3mm}
\change{
\subsection{Cross-Domain Interactions}
\label{subsec:cross}
A genuine unification of physical, sensor, and digital De-ID must articulate not only what each domain does in isolation, but how interventions in one domain interact with the others along the acquisition pipeline. 

First, a physical-domain perturbation must survive lens distortion, sensor noise, demosaicing, white balance, and downstream JPEG/H.264 compression before reaching a digital recognizer. Empirically, perturbations optimized without explicit modeling of the ISP pipeline (\emph{e.g.}, early adversarial patches) lose substantial attack success after realistic capture. Methods that explicitly model the capture chain (ProjAttacker~\cite{liu2025projattacker}, AT3D~\cite{yang2023towards}) close this gap, illustrating the necessity of co-design between physical perturbations and the sensing stage they must traverse.
Second, multi-stage interventions can compound or conflict. A sensor-domain phase mask combined with a downstream digital generative refiner (PrivacyOptics~\cite{lopez2024privacy}) leverages the strengths of both: the optics provide hardware-level identity suppression while the generator restores task utility. Conversely, a physical adversarial patch followed by a digital denoiser may be neutralized, since the patch is treated as removable noise.
Third, an intervention at an earlier stage can obviate the need for one at a later stage. Extreme low-resolution capture (ISR~\cite{ryoo2017privacy}) makes downstream digital pixelation redundant; conversely, strong digital generative De-ID (Swapping-DeID~\cite{kung2025face}) makes physical perturbations unnecessary when full digital pipeline control exists.
Fourth, the three domains differ fundamentally in where raw identity-bearing signal first exists in digital form. Physical-domain De-ID prevents identifiable signal from ever being produced; sensor-domain De-ID ensures that identifiable signal never leaves the camera; digital-domain De-ID accepts a fully identifiable signal and removes identity in software layer. The choice of domain is therefore primarily a choice of trust model rather than a choice of algorithm.}

\section{Face De-ID Methodologies}
\label{sec:method}

\begin{figure}[t]
    \centering
    \includegraphics[width=1.0\linewidth]{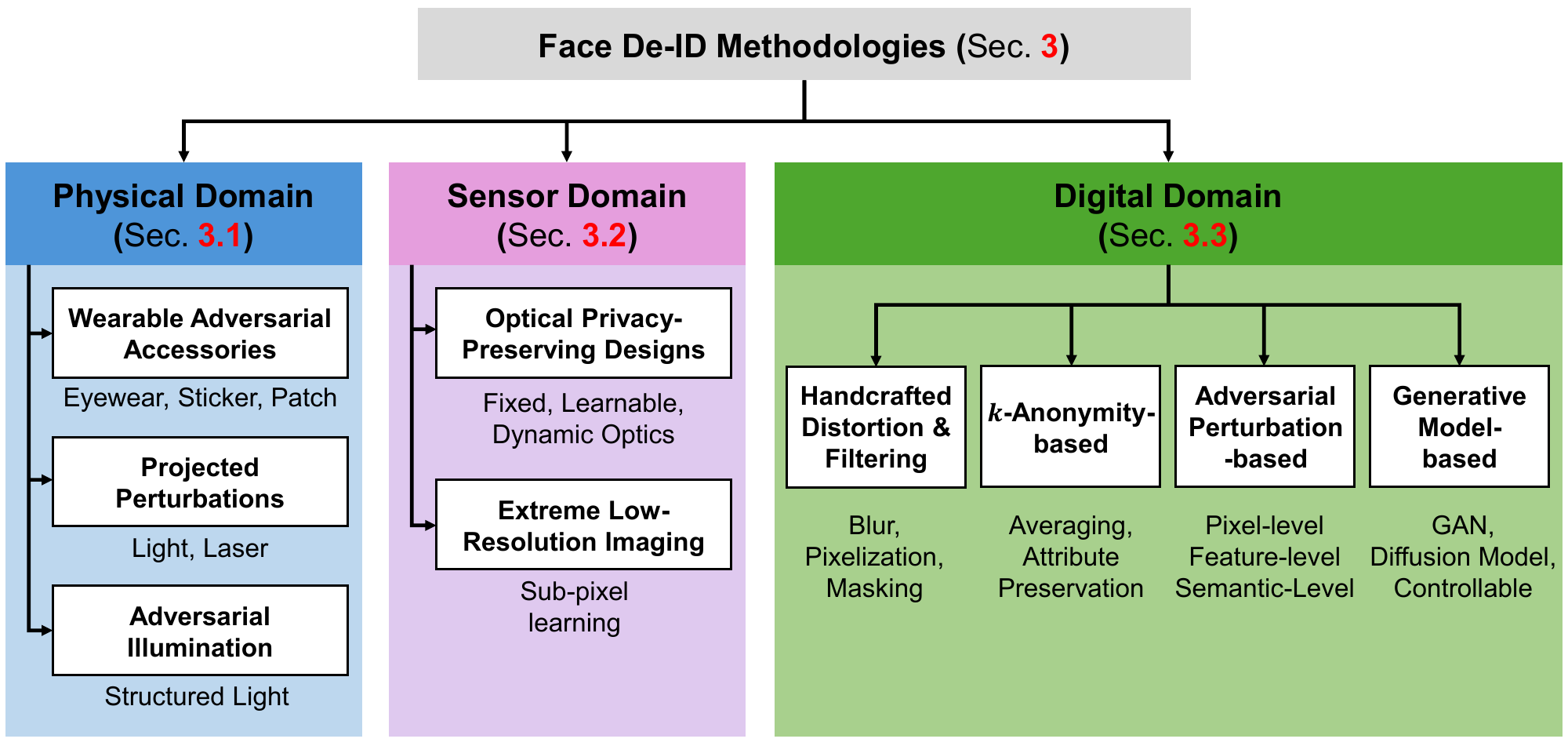}
    \vspace{-5mm}
    \caption{\textbf{Hierarchical taxonomy of face De-ID methodologies}. Each domain is further categorized into specific sub-classes based on their technical approaches.}
    \label{fig:method}
    \vspace{-5mm}
\end{figure}

Building upon the domain-centric taxonomy established in Sec.~\ref{subsec:Taxonomy}, we now present a comprehensive analysis of face De-ID methodologies across the physical, sensor, and digital domains.
Fig.~\ref{fig:method} provides a visual overview of the face De-ID methods discussed in this section.

\vspace{-3mm}
\subsection{Physical-Domain Face De-ID}
\label{sec:physical}

Physical-domain face De-ID manipulates the physical appearance of individuals or their surrounding environment to prevent FR systems from correctly identifying them. \change{Following Sec.~\ref{subsec:cross}, these methods intervene at the earliest stage of the acquisition pipeline: identity cues are suppressed on the subject before any digital signal exists, but the perturbation must subsequently survive optics, sensor, and ISP processing to remain effective at the recognizer.} Based on their mechanism, existing methods fall into three classes: \textbf{wearable adversarial accessories} (printed or fabricated items such as eyeglasses, hats, stickers, masks, and bandages), \textbf{projected perturbations} (external devices that cast adversarial patterns onto faces using visible or infrared light), and \textbf{adversarial illumination} (manipulated environmental lighting that creates naturally-appearing yet recognition-disrupting effects). Fig.~\ref{table:physical} illustrates representative examples.

\begin{table*}[t]
\centering
\renewcommand{\arraystretch}{1.2}
\setlength{\tabcolsep}{3pt}
\captionsetup{type=figure}
\resizebox{\textwidth}{!}{%
\begin{tabular}{ccccccccc}
\includegraphics[height=20mm]{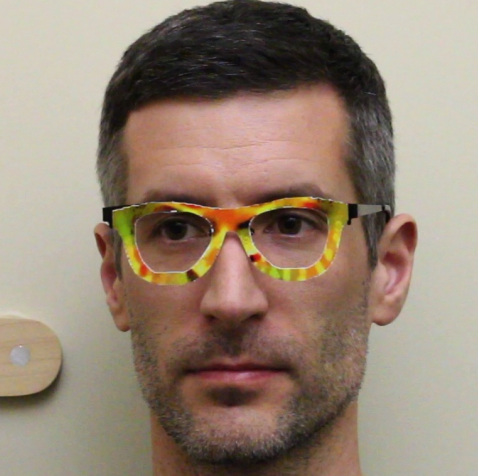} &
\includegraphics[height=20mm]{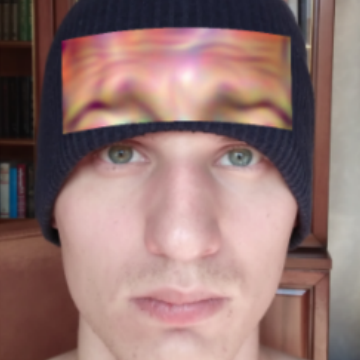} &
\includegraphics[height=20mm]{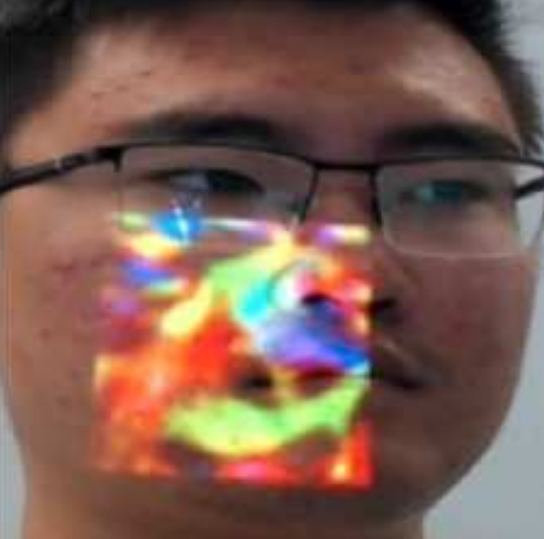} &
\includegraphics[height=20mm]{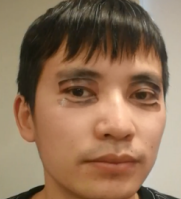} &
\includegraphics[height=20mm]{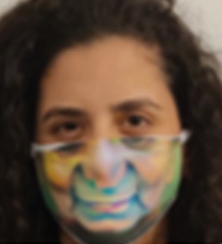} &
\includegraphics[height=20mm]{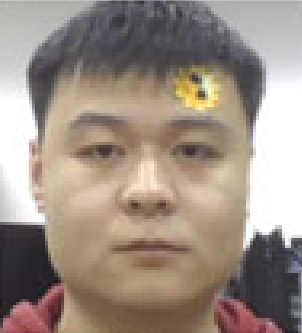} &
\includegraphics[height=20mm]{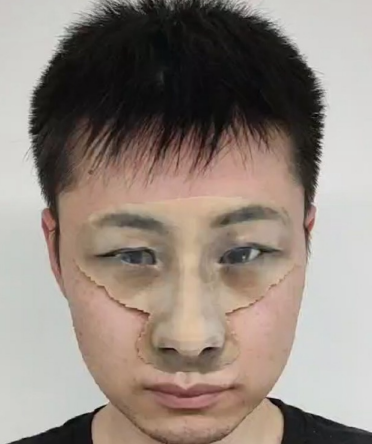} &
\includegraphics[height=20mm]{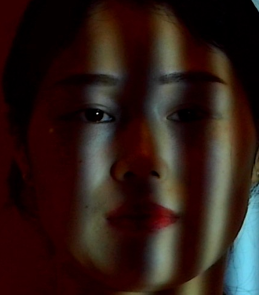} &
\includegraphics[height=20mm]{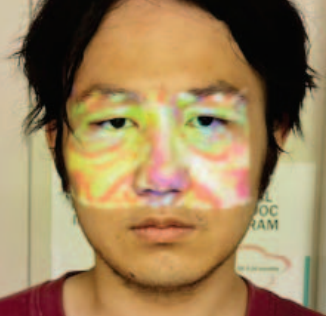} \\

\scriptsize (a) AdvEyeglass~\cite{sharif2016accessorize} &
\scriptsize (b) AdvHat~\cite{komkov2021advhat} &
\scriptsize (c) ALPA~\cite{nguyen2020adversarial} &
\scriptsize (d) AdvMakeup~\cite{yin2021adv} &
\scriptsize (e) AdvMask~\cite{zolfi2022adversarial} &
\scriptsize (f) AdvSticker~\cite{wei2022adversarial} &
\scriptsize (g) AT3D~\cite{yang2023towards} &
\scriptsize (h) Optical De-ID~\cite{li2023physical} & 
\scriptsize (i) ProjAttacker~\cite{liu2025projattacker} \\

\scriptsize  2016 &
\scriptsize 2020 &
\scriptsize 2020 &
\scriptsize  2021 &
\scriptsize 2022 &
\scriptsize  2022 &
\scriptsize 2023 &
\scriptsize 2023 &
\scriptsize  2025 \\

\end{tabular}
}
\vspace{-1mm}
\caption{\change{\textbf{Sample figures of representative physical‑domain face De-ID methods}. Wearable adversarial accessories: (a), (b), (d), (e), (f), (g). Projected perturbations: (c), (i). Adversarial illumination: (h).}}
\label{table:physical}
\vspace{-5mm}
\end{table*}

\change{Unlike digital or sensor methods, physical-domain face De-ID is governed by four dimensions that shape every design choice in this section. \emph{(1) Manufacturability and configurability}: the perturbation must be physically realizable using printing, fabrication, or projection hardware with bounded resolution, color gamut, and material reflectance. \emph{(2) Real-world robustness}: because the perturbation cannot be re-tuned at deployment time, it must remain effective across viewpoint, distance, motion, illumination, weather, background, and partial occlusion. \emph{(3) Conspicuity and social acceptability}: the perturbation is worn or projected in public, so its visual impact and social cost (legality, suspicion, attention) directly determine deployability, independently of attack success rate. \emph{(4) Propagation through sensor capture}: the perturbation passes through optics, sensor noise, ISP processing, and compression before reaching the recognizer, all of which can attenuate or destroy the adversarial signal. The following subsections examine the three classes of methods along these four axes.}

\subsubsection{Wearable Adversarial Accessories}
\textbf{Eyeglass-Based Approaches.}
Sharif \textit{et al.}~\cite{sharif2016accessorize} (AdvEyeglass) established the foundational paradigm for physical attacks against FR, introducing eyeglass frames with optimized textures that enable evasion or impersonation. Building on this, AGNs~\cite{sharif2019general} introduced a GAN-based framework that learns to generate adversarial eyeglasses from real eyeglass datasets, addressing vaguely specified objectives such as inconspicuousness that resist mathematical modeling.

\noindent\textbf{Patch and Sticker-Based Methods.}
The small perturbation area imposed by $L_{0}$-norm restrictions on eyeglasses motivates larger patches and stickers. AdvHat~\cite{komkov2021advhat} affixes rectangular stickers to hats for larger perturbation areas while preserving natural appearance. OAP~\cite{pautov2019adversarial} proposes grayscale patches for facial regions or eyeglasses, with a reproducible pipeline accounting for printing and camera color discrepancies. 
Physical AX~\cite{singh2022powerful} contributes a threshold-based smoothness loss with delayed total-variation penalties that improves convergence, and a patch-noise combo attack combining localized patches with imperceptible full-face noise for higher white-box and black-box success.

A central challenge is transferability in query-free black-box settings against unknown commercial systems. GenAP~\cite{xiao2021improving} regularizes patches on low-dimensional manifolds learned by pre-trained generative models (ProGAN~\cite{karras2018progressive}, StyleGAN~\cite{karras2019style,karras2020analyzing}), constraining them toward natural face features and narrowing the substitute-target response gap. PadvFace~\cite{zheng2023robust} models physical-world variations including chromatic aberration, sticker deformation, and illumination changes. EAP~\cite{liu2024eap} targets impersonation via random similarity transformations, image pyramids, and meta-ensemble attacks for robustness across scales, resolutions, and models. Other work moves beyond optimized patterns: AdvSticker~\cite{wei2022adversarial} uses real-life meaningful stickers (\emph{e.g.}, decorative designs); PPAttack~\cite{wei2023simultaneously} jointly optimizes patch perturbation and position via reinforcement learning; and FaceAdv~\cite{shen2021effective} uses Grad-CAM~\cite{selvaraju2017grad} to locate critical regions, with a generator-converter architecture simulating physical capture under limited sticker area.

\noindent\textbf{Makeup and Mask-Based Approaches.}
Cosmetics offer a vehicle for inconspicuous perturbations. AdvMakeup~\cite{yin2021adv} pioneered this with a GAN-based approach synthesizing natural adversarial eye shadow over the orbital region, combining gradient constraints with VGG16-based style/content losses~\cite{simonyan2015very} for imperceptibility. The COVID-19 normalization of face masks enabled mask-based methods. Adversarial Mask~\cite{zolfi2022adversarial} embeds Universal Adversarial Perturbations into fabric masks through differentiable digital masking that places perturbations precisely along facial contours. SASMask~\cite{gong2023stealthy} instead optimizes adversarial mask styles via continuous relaxation, maximizing impersonation success while staying visually realistic across digital, physical, and commercial platforms. AdvBandage~\cite{bhilare2024fooling} uses perturbed medical bandages as discreet patches, optimizing bandage size and placement with iterative FGSM~\cite{GoodfellowSS14} for both dodging and impersonation, with bandages providing natural cover without arousing suspicion.

\noindent\textbf{3D Adversarial Meshes.}
2D accessories are limited by detectability under anti-spoofing defenses and reduced effectiveness against commercial systems. 3D adversarial meshes address this by modeling the full geometry of facial perturbations for robust attacks across viewing conditions. AT3D~\cite{yang2023towards} pioneered this with Adversarial Textured 3D Meshes that are 3D-printed and applied directly to faces; its key innovation is optimization in the low-dimensional coefficient space of 3D Morphable Models (3DMM)~\cite{tuan2017regressing} rather than high-dimensional mesh space, which accelerates optimization, avoids local optima, and improves black-box transferability. Physical tests showed evasion of four commercial anti-spoofing APIs, two mobile systems, and two access control systems. Face3DAdv~\cite{yang2025face3dadv} extends this with a controllable simulation framework that reconstructs full 3D face information (texture, shape, viewpoint, lighting) via a differentiable renderer, using importance sampling to prioritize critical physical transformations.

\vspace{-1mm}
\subsubsection{Projected Perturbations}
Projection-based methods shift from accessories to artifact-free De-ID by manipulating light patterns cast onto faces, offering real-time adaptability and multi-target flexibility.

ALPA~\cite{nguyen2020adversarial} pioneered this direction with real-time adversarial light projections using an off-the-shelf camera-projector setup, incorporating landmark-based position calibration and Lab-space color calibration to accurately reproduce adversarial patterns, while generating transformation-invariant patterns through representative face averages. ProjAttacker~\cite{liu2025projattacker} addressed digital-physical gap challenges through 3DMM~\cite{tuan2017regressing} initialization for geometry alignment, Light Reflection Function modeling for interactions among projected light, skin reflectance, and ambient illumination, and a differentiable camera ISP proxy network simulating real-world imaging variations. Agile~\cite{wang2024invisible} pushes stealthiness further with adjustable, invisible infrared laser emissions directed into camera CMOS sensors rather than visible-light projection, enabling Denial-of-Service, dodging, and impersonation attacks.

\vspace{-1mm}
\subsubsection{Adversarial Illumination}
Unlike projection-based methods that actively cast patterns onto faces, adversarial illumination exploits FR sensitivity to lighting by manipulating environmental lighting or the illumination used in specialized 3D acquisition systems, achieving De-ID through naturally-appearing lighting variations that avoid the conspicuousness of projected patterns.

Optical De-ID~\cite{li2023physical} targets structured-light-based 3D FR by corrupting the 3D data acquisition process itself, integrating 3D reconstruction and skin reflectance models into optimization through two strategies: phase shifting attacks that modify projected structured light patterns, and phase superposition attacks that project external adversarial noise. The dual approach enables adversarial point placement anywhere on the face with robustness to head movements. ARA~\cite{zhang2024adversarial} exploits illumination vulnerabilities in 2D FR by producing naturally relighted face images, combining physical model-based optimization (which guides adversarial lighting using feedback from FR systems) with efficient neural-network-based prediction (instantaneous prediction of adversarial light settings), with physical-world verification using precise relighting hardware.

\vspace{-1mm}
\subsubsection{Discussion}
\change{Physical-domain face De-ID has progressed from handcrafted, accessory-bounded textures to 3D- and optics-aware pipelines that explicitly model capture physics and sensor behavior. Reviewing this progress along the four physical-domain dimensions, several gaps persist. On \emph{manufacturability}, low-dimensional priors such as GAN manifolds~\cite{goodfellow2014generative}, 3DMM coefficients~\cite{tuan2017regressing}, and style spaces~\cite{xiao2021improving,gong2023stealthy} have systematically improved transfer and realism by constraining perturbations near the face manifold, but standardized fabrication tolerances and material-aware optimization remain underdeveloped, particularly for 3D-printed meshes~\cite{yang2023towards,yang2025face3dadv} where print resolution and material reflectance directly affect deployment fidelity. On \emph{real-world robustness}, non-contact photonic channels (visible/IR projection~\cite{nguyen2020adversarial,liu2025projattacker,wang2024invisible}, structured-light interference~\cite{li2023physical}) widen the attack surface while alleviating occlusion-induced liveness failures, yet standardized benchmarks encompassing multi-view video, multi-sensor (RGB-IR-depth) fusion, and liveness under varied ambient conditions are still lacking, as are cross-system generalization studies under policy and threshold changes (including open-set verification) and principled evaluations of long-term stability such as washability, wear, and battery/thermal constraints for emitters~\cite{nguyen2020adversarial,wang2024invisible}. On \emph{conspicuity and social acceptability}, quantitative stealth metrics aligned with human perception and social norms are needed; current proxies based on area or total-variation~\cite{singh2022powerful} under-capture social cost, and even methods designed for inconspicuousness~\cite{yin2021adv,zolfi2022adversarial,bhilare2024fooling} rely largely on qualitative argument rather than measurement. On \emph{propagation through sensor capture}, methods that explicitly model the capture chain~\cite{liu2025projattacker,yang2023towards} demonstrate the value of physical--sensor co-design, but co-design with defenses that jointly perform spoof detection, quality assessment, and identity inference remains largely unexplored. Addressing these gaps will require reproducible, physics-grounded testbeds, shared calibration protocols, and reporting standards that elevate claims from single-device demonstrations to robust, deployable face De-ID guarantees.}

\vspace{-3mm}
\subsection{Sensor-Domain Face De-ID}
\label{sec:sensor}

Sensor-domain face De-ID methods intervene during image acquisition, preventing identifiable facial information from ever being recorded.
\change{Following the cross-domain analysis of Sec.~\ref{subsec:cross}, sensor-domain methods intervene inside the camera: identity-bearing signal exists optically but is encoded before digitization, and these methods frequently combine with downstream digital generators that restore task utility from the privacy-encoded measurement.}
These techniques fall into two paradigms: \textbf{optical privacy-preserving designs} that manipulate incident light through defocus, aberrations, or phase masks before it reaches the sensor, and \textbf{extreme low-resolution imaging} that captures faces below the recognition threshold. \change{Fig.~\ref{fig:sensor} contrasts the two paradigms: (a) shows the optical pipeline, in which a point source is encoded by the camera optics into a privacy-preserving point spread function (PSF) before reaching the sensor, across its fixed, learnable, and dynamic generations; (b) shows the low-resolution pipeline, in which a tiny camera records faces too coarse for recognition while a downstream network still recovers task utility.}

\vspace{-1mm}
\subsubsection{Optical Privacy-Preserving Designs}

\begin{figure}[t]
    \centering
    \includegraphics[width=1.0\linewidth]{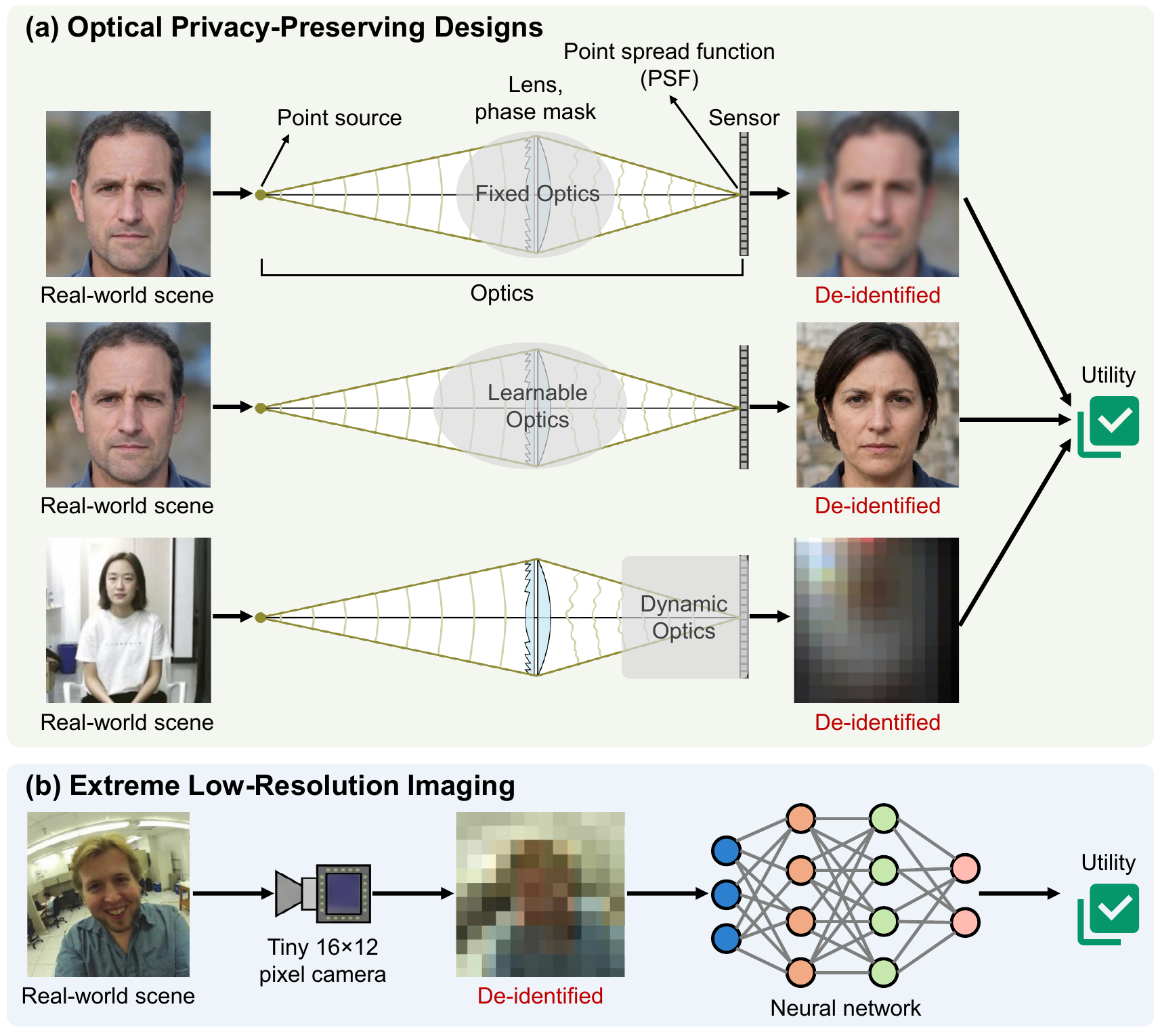}
    \caption{\change{\textbf{Taxonomy of sensor-domain face De-ID paradigms.} \textbf{(a)~Optical privacy-preserving designs} encode privacy in the camera optics: a point source is shaped by a lens and phase mask into a point spread function (PSF) recorded by the sensor, yielding a de-identified image while downstream utility is retained. Three generations are shown, namely \emph{fixed} optics with hand-specified degradation (DefocusOptics~\cite{pittaluga2015privacy}, DefocusOptics$^+$~\cite{pittaluga2016pre}), \emph{learnable} optics jointly optimized with downstream networks (PrivHPE~\cite{hinojosa2021learning}, PrivHAR~\cite{hinojosa2022privhar}, PrivPDE~\cite{tasneem2022learning}, OpticalDR~\cite{pan2024opticaldr}, PrivacyOptics~\cite{lopez2024privacy}), and \emph{dynamic} PSF that randomize the configuration per capture (DyPP~\cite{cheng2024learning}). \textbf{(b)~Extreme low-resolution imaging} uses a tiny camera (e.g., 16$\times$12 pixels) to capture faces below the recognition threshold, with a downstream neural network recovering task utility (ISR~\cite{ryoo2017privacy}, LRPrivacy~\cite{ryoo2018extreme}).}}
    \label{fig:sensor}
    \vspace{-4mm}
\end{figure}

\change{Optical designs encode privacy in the imaging optics themselves. As illustrated in Fig.~\ref{fig:sensor}(a), they have progressed through three generations, namely \emph{fixed} designs with hand-specified degradation, \emph{learnable} designs that jointly optimize optics and downstream networks, and \emph{dynamic} designs that vary the optical configuration at capture time.}

\noindent\textbf{\change{Fixed Optics.}}
\change{Early optical privacy methods employed fixed designs to attenuate facial features.} DefocusOptics~\cite{pittaluga2015privacy} pioneered $k$-anonymity~\cite{sweeney2002k} preserving optical designs using intentional defocus blur, providing foundations for miniaturizing privacy optics within small sensor volumes. DefocusOptics$^+$~\cite{pittaluga2016pre} extended this with programmable optics enabling pre-capture mask-based techniques. \change{However, these hand-specified defocus mechanisms proved vulnerable to deconvolution attacks, motivating learnable optical designs with complex, irreversible degradations.}

\noindent\textbf{\change{Learnable Optics.}}
\change{Learnable optical approaches jointly optimize hardware and software components for task-specific privacy-utility trade-offs.} PrivHPE~\cite{hinojosa2021learning} introduced end-to-end optimization of optical encoders (parametrized via Zernike coefficients) and CNN decoders for pose estimation, using privacy-preserving loss functions to minimize face keypoint detection while maintaining task performance. PrivHAR~\cite{hinojosa2022privhar} advanced this through adversarial training of phase masks for action recognition, achieving near-random classification performance on FR while preserving action recognition accuracy. Further refinements emerged through adversarial learning frameworks: PrivPDE~\cite{tasneem2022learning} balanced FR prevention with depth estimation via aperture-plane phase masks, OpticalDR~\cite{pan2024opticaldr} achieved near-random FR while preserving depression-related features, and PrivacyOptics~\cite{lopez2024privacy} combined optical encoding with GAN-based face synthesis to close the security gap between capture and De-ID.

\noindent\textbf{\change{Dynamic Optics.}}
\change{A critical vulnerability of fixed and learnable-but-static designs is susceptibility to PSF inversion attacks, where adversaries recover the point spread function to reconstruct faces.} DyPP~\cite{cheng2024learning} addressed this by introducing time-varying PSFs sampled from a learned privacy manifold, implemented via spatial light modulators. This dynamic design randomizes the optical configuration for each capture, significantly enhancing robustness against inversion attacks while maintaining task utility for object detection and pose estimation.

\vspace{-1mm}
\subsubsection{Extreme Low-Resolution Imaging}
\change{Rather than encoding privacy in the optics, extreme low-resolution imaging achieves face De-ID by capturing at resolutions} (e.g., 16$\times$12 pixels) where facial detail is inherently insufficient for reliable recognition (faces are reduced to 2$\times$2 or smaller representations). \change{As shown in Fig.~\ref{fig:sensor}(b), the coarse capture is fed directly to a downstream network that recovers task utility without ever recording an identifiable image.} ISR~\cite{ryoo2017privacy} pioneered this paradigm by proposing inverse super resolution, generating multiple informative low-resolution training videos from high-resolution sources via optimized sub-pixel transformations. This enables learning robust decision boundaries in low-resolution feature space while ensuring hardware-level privacy, achieving comparative activity recognition accuracy. LRPrivacy~\cite{ryoo2018extreme} advanced this through a two-stream multi-Siamese CNN that learns transformation-invariant embeddings, enabling real-time activity classification on mobile GPUs without recording identifiable high-resolution videos.

\vspace{-1mm}
\subsubsection{Discussion}
Sensor-based face De-ID offers strong security-by-design advantages by eliminating the raw identifiable face from the acquisition path, but several challenges persist. First, threat models and evaluation protocols require standardization: claims of irreversibility should be stress-tested against optical inversion, deconvolution/deblurring, super-resolution, face restoration, and ISP-level side channels across diverse sensors and color pipelines. Second, dynamic optics (e.g., DyPP) improve resilience but introduce calibration, latency, and energy costs; quantifying these trade-offs for mobile and embedded platforms is essential. Third, privacy--utility objectives remain task- and population-dependent: preserving low-frequency geometry can leak soft-biometrics, and fairness across demographics under optical distortions is underexplored. Fourth, co-design with downstream ISPs and multi-camera systems is largely unaddressed, as is long-term robustness to hardware drift and manufacturing tolerances. 
Finally, reproducible cross-device benchmarks coupling privacy, task, and perception metrics are needed for on-sensor De-ID at scale.

\begin{figure*}[t]
\centering    
\subfigure[Handcrafted methods: Blur, pixelization, masking, and filtering.]{\label{fig:early1}\includegraphics[height=24.0mm]{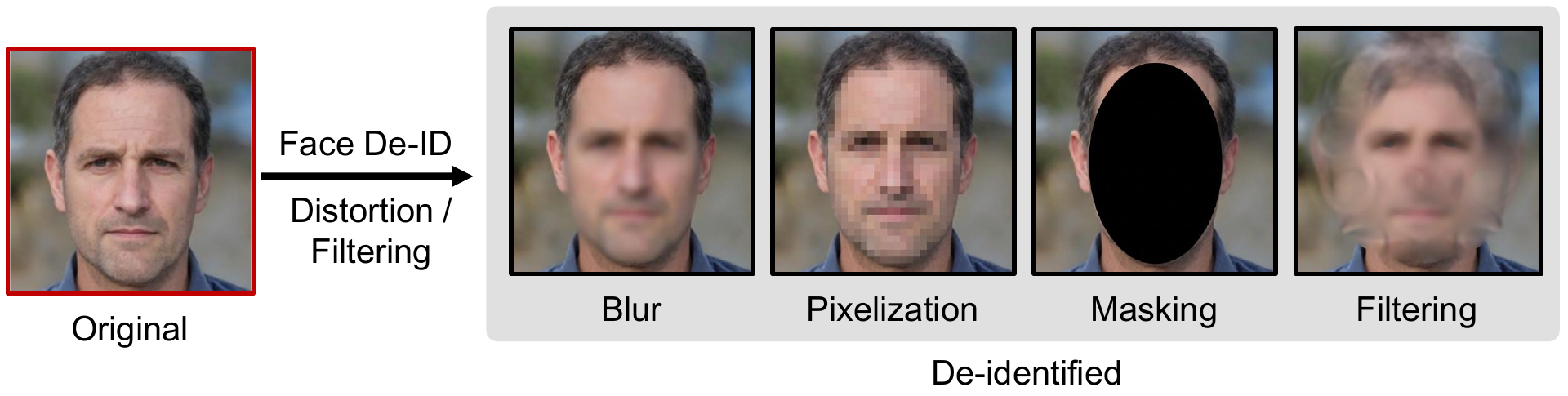}} \hspace{2mm}
\subfigure[$k$-Same method: Cluster $k$ faces, substitute with aggregate.]{\label{fig:early2}\includegraphics[height=24.0mm]{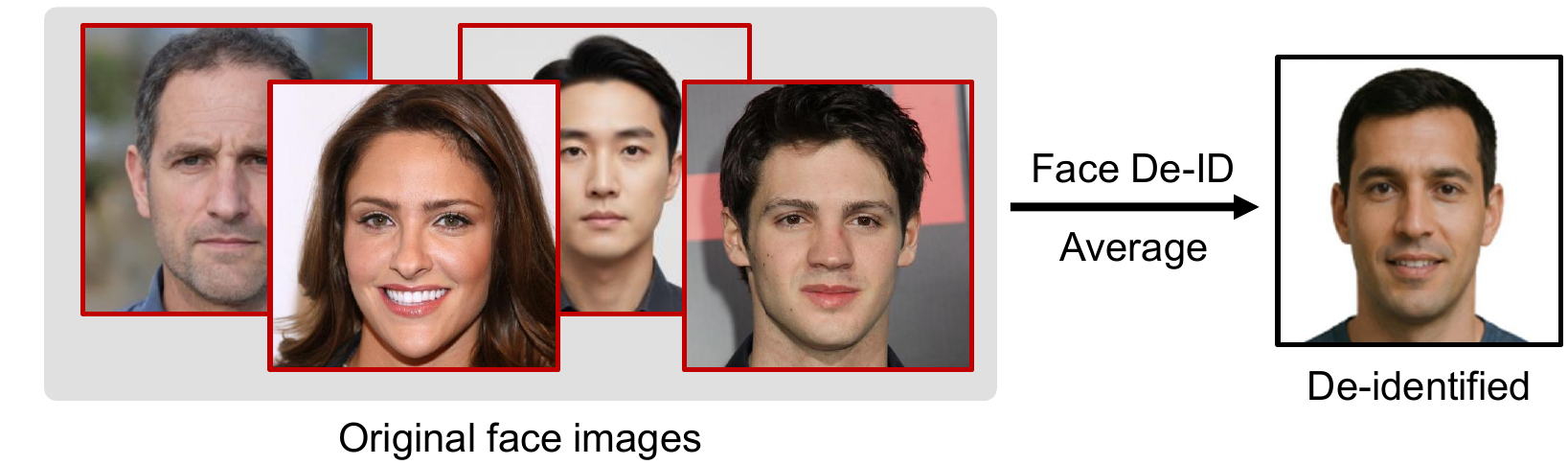}}
\vspace{-2mm}
\caption{\textbf{Traditional digital-domain face De-ID approaches.} (a) Handcrafted filters, including blur, pixelization, and masking, uniformly reduce the recognizability of the facial region, alongside spatially adaptive filtering (EmotionPreserve~\cite{letournel2015face}) that selectively blurs identity-bearing structures while preserving utility-bearing regions. (b) The $k$-Same framework aggregates a cluster of $k$ face images to synthesize a mean face that replaces the original, theoretically limiting the recognition probability to $1/k$. Together they illustrate the shift from ad-hoc distortion to statistically grounded face De-ID.}
\label{fig:early}
  \vspace{-0.4cm}
\end{figure*}

\vspace{-3mm}
\subsection{Digital-Domain Face De-ID}
\label{sec:digital}

\change{Following the cross-domain analysis of Sec.~\ref{subsec:cross}, digital-domain methods accept a fully identifiable digital signal and suppress identity in software. Their flexibility comes at the cost of accepting raw exposure and, in some regimes, replacing rather than complementing upstream physical or sensor interventions.}

\subsubsection{\change{Handcrafted Distortion and Filtering}}
\label{sec:naive}
Handcrafted distortion methods are the earliest face De-ID approaches, employing simple image processing operations. Fig.~\ref{fig:early}(a) shows naive distortion (Gaussian blur, pixelization, masking), while Fig.~\ref{fig:early}(b) depicts the $k$-Same framework, marking the shift from ad-hoc distortion to statistically grounded face De-ID.

Early work evaluated these operations for their privacy-utility trade-offs. EFVAP~\cite{hudson1996techniques} and De-ID Filter~\cite{boyle2000effects} studied blur and pixelization in workplace video, finding that moderate blur conceals identity while preserving awareness cues such as posture, activity, and people count. Blur outperformed pixelization, which showed abrupt jumps in information disclosure across levels. However, BFPP~\cite{neustaedter2006blur} exposed a limitation in home telecommuting: blur weak enough to preserve awareness cues failed to de-identify, since motion and context still enabled recognition. This tension motivated spatially adaptive techniques that selectively preserve task-relevant information. PUI~\cite{crowley2000things} introduced the Eigen-space filter, projecting live images onto a PCA basis learned from ``socially correct'' reference images so that reconstruction retains only reference-set information while suppressing non-conforming elements, though this raised ethical concerns over misuse for face animation and masquerading. EmotionPreserve~\cite{letournel2015face} instead preserved expressions via variational adaptive filtering with Total Variation regularization, strongly blurring identity-contributing structures (nose, wrinkles) while sparing expression-related regions (eyes, eyebrows, mouth).

\textbf{Discussion.}
Handcrafted distortion established the foundational framework for digital face De-ID but suffers from critical limitations. Uniform operations cannot achieve strong De-ID while preserving utility, owing to the coupling between identity and task-relevant features: privacy-preserving settings eliminate awareness cues, while utility-preserving settings leave enough signal for recognition through motion and context. These methods also lack formal privacy guarantees, relying on user studies or model-specific recognition accuracy rather than rigorous metrics. Their need for empirical, context-dependent tuning shows that handcrafted distortion cannot universally balance privacy and utility, motivating learning-based approaches that adaptively optimize this trade-off.

\subsubsection{$k$-Anonymity-Based Methods}
\label{sec:ksame}

Formal privacy notions migrated into face De-ID via $k$-anonymity~\cite{sweeney2002k}, which bounds re-identification risk by ensuring each released record is indistinguishable from at least $k-1$ others. The foundational $k$-Same algorithm~\cite{newton2005preserving} translates this principle to faces by clustering images and replacing each with the cluster average, guaranteeing recognition accuracy $\leq 1/k$. Early refinements addressed utility and quality: $k$-Same-Select~\cite{gross2005integrating} imposed utility constraints to preserve gender and expression during clustering, while $k$-Same-M~\cite{gross2006model} leveraged active appearance models for improved alignment and photometric fidelity. These works systematically exposed weaknesses in ad-hoc pixelation, including susceptibility to parrot recognition and resolution enhancement attacks~\cite{gross2006model}.

Subsequent developments expanded beyond closed-set scenarios. The $\epsilon$-map framework~\cite{gross2007towards} formalized three privacy targets: $\epsilon$-map (``like no one''), wrong-map (``like someone else''), and $(\epsilon,k)$-map (``like everyone''), via adaptive pixelation and likelihood equalization under reference models, supporting multiple images per subject with explicit privacy bounds. Multi-Factor DeID~\cite{gross2008semi} disentangled identity from nuisance factors using unified linear/bilinear/quadratic models with semi-supervised fitting. An alternative emerged with $k$-Same-furthest~\cite{meng2014face}, which averaged over maximally dissimilar clusters rather than similar ones, theoretically driving recognition toward zero while maintaining $k$-anonymity.

Attribute preservation became central in later methods. GARP-Face~\cite{du2014garp} blended faces with attribute-matched gallery images to preserve demographics while removing identity. APFD~\cite{jourabloo2015attribute} formulated this as joint optimization over AAM parameters, computing optimal fusion weights for $k$ attribute-similar images. 
\change{Deep learning enabled new approaches: FIP-DeID~\cite{chi2015face} exploited pose-invariant features to generate averaged faces; $k$-Same-Net~\cite{meden2018k} replaced averaging with a generator trained on a disjoint proxy identity set, eliminating pixel-averaging artifacts while preserving the $1/k$ bound; ELEGANT-DeID~\cite{yan2019attributes} combined $k$-Same with generative attribute transfer, achieving complete preservation with minimal gallery requirements.}

\textbf{Discussion.} $k$-Anonymity methods established the first formal privacy framework for face de-identification, providing provable guarantees that recognition accuracy cannot exceed $1/k$. However, this guarantee entails fundamental limitations: clustering constraints reduce scalability and introduce bias with skewed distributions, while pixel averaging produces perceptual artifacts. More critically, $k$-anonymity addresses only closed-world threat models, offering no protection against adversaries with auxiliary information. The utility–privacy trade-off remains coarse-grained (reducing $k$ to preserve visual detail discontinuously increases re-identification risk) and attribute-preservation extensions~\cite{jourabloo2015attribute,yan2019attributes} require extensive labeled data with limited control over attribute retention. Despite these constraints, $k$-anonymity methods laid essential conceptual groundwork for subsequent developments.

\begin{figure*}[t]
    \centering
    \includegraphics[width=1.0\linewidth]{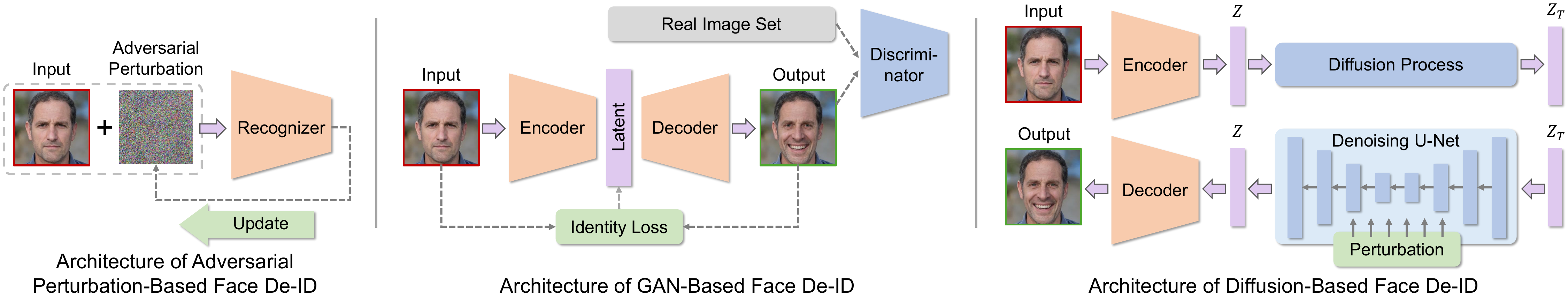}
    \caption{\textbf{Architectural paradigms of contemporary digital-domain face De-ID}. (Left) Adversarial perturbation-based methods inject imperceptible noise at the pixel level to mislead face recognizers. (Middle) GAN-based generative approaches synthesize identity-removed faces through encoder–decoder frameworks guided by discriminators and identity losses. (Right) Diffusion-based methods employ iterative denoising processes to achieve high-fidelity, photorealistic anonymization within structured latent spaces. Together, these architectures illustrate the evolution from explicit perturbation to implicit generative synthesis for controllable, realistic, and utility-preserving face De-ID.}
    \label{fig:generative}
    \vspace{-4mm}
\end{figure*}

\subsubsection{Adversarial Perturbation-based Methods}
\label{sec:adversarial}

Early adversarial perturbation approaches generated minimal, imperceptible noise to fool FR systems (Fig.~\ref{fig:generative}, left). EvolutionaryAttack~\cite{dong2019efficient} showed that perturbations as small as $10^{-5}$ MSE enable dodging attacks, while P-FGVM~\cite{chatzikyriakidis2019adversarial} added realism constraints to spatial-domain perturbations, and AdvFaces~\cite{deb2020advfaces} used discriminators for high perceptual quality under both dodging and impersonation. These pixel-level approaches, though effective, struggled with visual naturalness and cross-system transferability.

To address this, researchers turned to feature-space manipulation in learned representations rather than pixel space. FSAP~\cite{xue2023face} optimizes latent vectors via bi-loss alternation between identity dissimilarity and attribute preservation; TIP-IM~\cite{yang2021towards} combines relative identification loss with Maximum Mean Discrepancy constraints for open-set naturalness; FE-DeID~\cite{hanawa2024face} concentrates perturbations on critical regions (eyes, nose, mouth) via feature embedding; and Adv-Inversion~\cite{wang2025adv} optimizes latent codes through reconstruction-fidelity losses, improving transferability over pixel-level perturbations.

Other methods use natural visual modifications as perturbation carriers for stealthiness. AMT-GAN~\cite{hu2022protecting} reconciles adversarial noise with cycle consistency in makeup transfer through joint training, while Makeup De-ID~\cite{zhu2019generating} embeds perturbations in eye-makeup regions for more realistic dodging and impersonation. CLIP2Protect~\cite{shamshad2023clip2protect} searches latent manifolds guided by textual makeup prompts, and DiffProtect~\cite{liu2023diffprotect} produces target-agnostic examples that avoid noise-based artifacts, while WeakenDiff~\cite{salar2025enhancing} targets purification effects via learned unconditional embeddings to raise impersonation success.

Semantic-level manipulation has gained attention for its natural appearance and transferability. Adv-Attribute~\cite{NEURIPS2022_dccbeb7a} perturbs high-level attributes in disentangled latent spaces through importance-aware selection and multi-objective optimization, excelling against robust black-box models. Adv-CPG~\cite{wang2025advcpg} integrates customized portrait generation with progressive two-layer identity encryption, enabling multi-modal control while achieving strong black-box success.

Black-box transferability remains a central challenge. DFANet~\cite{zhong2020towards} uses dropout to diversify examples and prevent surrogate overfitting, establishing the TALFW robustness benchmark; Sibling-Attack~\cite{li2023sibling} adds attribute recognition as an auxiliary task to improve generalization; SMAP~\cite{ma2023transferable} jointly optimizes patch texture, position, and shape via gradient-guided location selection; and ADA~\cite{hu2024toward} and DPA~\cite{zhou2025improving} respectively survey transferability enhancement across FR architectures and advance unrestricted-attack embedding strategies.

A further group targets specialized scenarios. GMAA~\cite{li2023discrete} produces expression-robust impersonation by training across facial action units; UAXs~\cite{amada2021universal} exposes vulnerabilities via identity-agnostic universal perturbations that spoof multiple identities at once; VLA~\cite{shen2019vla} extends attacks to physical settings using projected visible light with alternating perturbation and concealing frames; Chameleon~\cite{chow2024personalized} offers instant protection through identity-preserving targeted perturbations; and Veil Privacy~\cite{pang2024veil} generates veiled data that conceals identity from humans while preserving, or enhancing, DNN utility.

\textbf{Discussion}.
Adversarial perturbation methods have evolved from pixel-level noise injection~\cite{dong2019efficient,chatzikyriakidis2019adversarial} through feature-space methods~\cite{xue2023face,yang2021towards} to semantic manipulations~\cite{NEURIPS2022_dccbeb7a,wang2025adv}, a clear trend toward more natural-looking and transferable perturbations. Yet four challenges persist. First, the \textit{transferability-imperceptibility trade-off}: perturbations effective on surrogates often fail on black-box systems, while larger magnitudes compromise naturalness~\cite{zhong2020towards,li2023sibling}. Second, limited \textit{robustness to preprocessing} (JPEG compression, resizing, filtering), which attenuates adversarial effects before reaching the target. Third, variable \textit{computational efficiency}, with iterative optimization~\cite{yang2021towards,xue2023face} limiting real-time use. Fourth, \textit{evaluation inconsistency} across FR models, datasets, and metrics, which hinders systematic comparison. Future work should prioritize unified protocols assessing not only attack success but also visual quality, transferability, and efficiency.

\subsubsection{Generative Model-based Methods}
\label{sec:gennerative}

GANs have emerged as a dominant paradigm for face De-ID, enabling the synthesis of photorealistic faces while preserving non-identity attributes (Fig.~\ref{fig:generative}, middle). Early approaches established foundational architectures through style transfer and conditional generation. FATM~\cite{li2019identification} pioneered facial attribute transfer through encoder-decoder architectures mapping non-identity attributes to donor identities. EPD-Net~\cite{aggarwal2020epd} introduced dual auxiliary networks with identity and emotion verificators to maximize emotion preservation while minimizing identity similarity. Live-DeID~\cite{gafni2019live} extended these principles to video streams through attractor-repeller mechanisms where low-to-mid level features enforce similarity to input frames while high-level representations enforce distance from target identities.

Several methods have addressed privacy-utility trade-offs through specialized architectures and loss formulations. PP-GAN~\cite{wu2019privacy} combined Siamese verificators for identity removal with SSIM-based regulators for utility preservation. Generative-DeID~\cite{brkic2017know} extended De-ID to full-body scenarios through segmentation-conditioned synthesis. AutoDeID~\cite{kuang2021effective} integrated $k$-anonymity principles through Anonymous Semantic Masks and identity-adversarial discriminators ensuring distance from original and sensitive identities. DeID-GAN~\cite{agarwal2021privacy} utilized StyleGAN-generated proxy faces as anonymous templates, fusing non-biometric attributes based on emotion and pose similarity.

Attribute-aware and controllable generation has become increasingly prominent. A$^3$GAN~\cite{zhai2022a3gan} formulated face De-ID as joint semantic suppression and controllable attribute injection through suppressive convolutional units and attribute-aware injective networks. FaceSwap~\cite{yang2021systematical} combined StyleGAN-based attribute disentanglement with adversarial vector mapping in latent space to generate recognition-resistant images. SF-GAN~\cite{li2021sfgan} distinguished between shallow attributes (hairstyle, glasses) and deep attributes (expression, gender) through specialized processing networks, employing uniqueness loss to ensure distinctness. AnonymousNet~\cite{li2019anonymousnet} achieved measurable privacy through four-stage frameworks integrating attribute selection compliant with $k$-anonymity, $l$-diversity, $t$-closeness, and $\varepsilon$-differential privacy. \change{Disguise~\cite{cai2024disguise} grounded De-ID in differential privacy and ensemble learning with mixture-of-experts utility networks for attribute disentanglement.}

Reversibility represents critical advances for practical deployment. IdentityMask~\cite{wen2022IdentityMask} introduced reversible video De-ID with Protection and Recovery Modules using user-specific keys and motion flow guidance for temporal consistency. PIDIM~\cite{cao2021personalized} enabled personalized invertible De-ID through password-controlled hyperspherical transformations. UU-Net~\cite{proencca2022uu} proposed landmarks-free reversible solutions using sequential encoder-decoder models for public anonymization and private reconstruction. IDeudemon~\cite{wen2023divide} adopted divide-and-conquer strategies obfuscating 3D disentangled identity codes from NeRF models while preserving utility through visual similarity assistance.

Contemporary methods leverage advanced architectures including StyleGAN variants and diffusion models. StyleGAN-based approaches demonstrate particular promise: StyleGAN-DeID~\cite{khorzooghi2023examining} achieved effective De-ID through style mixing preserving utility attributes; FALCO~\cite{barattin2023attribute} operated in StyleGAN2 latent space with margin-based identity obfuscation in ArcFace space and feature-matching attribute preservation in FaRL's ViT space, eliminating background-based re-identification risks; and CPP-DeID~\cite{meden2023face} enabled customizable privacy-utility trade-offs through GAN inversion with privacy parameter-controlled identity suppression in $W^+$ latent space.

\change{{Diffusion-based De-ID} (Fig.~\ref{fig:generative}, right) has emerged as a powerful alternative to GANs. DiffAM~\cite{sun2024diffam} introduced diffusion-based adversarial makeup with superior black-box transferability. Diff-Privacy~\cite{he2024diff} unified anonymization via Multi-Scale image Inversion with embedding scheduling. Synthetic-DeID~\cite{park2025facial} reformulated De-ID as training-free identity editing in frozen diffusion models, and Swapping-DeID~\cite{kung2025face} simplified the pipeline through reconstruction loss within Stable Diffusion, controlling De-ID via a single parameter.}

\change{{Multi-spectral and 3D-aware approaches} extend De-ID beyond RGB still imagery. Thermal-F De-ID~\cite{lin2020identification} combined RGB and thermal imagery with ensemble learning, showing that thermal features prevent photo-based spoofing. MVC-DeID~\cite{cao2023achieving} leveraged 3D-aware StyleNeRF for viewpoint-consistent identity disentanglement, while G$^2$Face~\cite{yang2024g} integrated 3D face models and StyleGAN decoders for password-based reversibility.}

\change{{Context-specific applications} are defined by application-driven utility constraints: DeID-rPPG~\cite{savic2023identification} preserves remote photoplethysmography signals; FaceMotionPreserve~\cite{zhu2024facemotionpreserve} maintains facial motion for medical diagnosis; Egocentric-DeID~\cite{puangthamawathanakun2023towards} targets wearable-camera footage; Veil Privacy~\cite{pang2024veil} balances human-imperceptible obfuscation with DNN utility; and DRGAN~\cite{liu20253d} pioneered 2D/3D De-ID preserving expression, gender, and ethnicity through disentanglement-reconstruction.}

\subsubsection{Discussion}
Generative methods represent the most sophisticated paradigm for face De-ID. 
The field has progressed from early conditional GANs prioritizing photorealism to frameworks incorporating formal privacy guarantees ($k$-anonymity~\cite{kuang2021effective}, differential privacy~\cite{li2019anonymousnet}), reversible mechanisms~\cite{cao2021personalized,wen2022IdentityMask}, and latent space manipulation~\cite{meden2023face,barattin2023attribute}. Despite superior visual quality, critical challenges remain: (1) the privacy-utility trade-off constitutes a fundamental limitation: perfect identity removal conflicts with perfect attribute preservation given representation entanglement; (2) evaluation fragmentation persists across disparate datasets and metrics, precluding meaningful comparison; (3) computational costs prohibit real-time deployment, particularly for diffusion approaches~\cite{kung2025face}; and (4) temporal consistency for video receives insufficient attention, with most methods designed for images. Recent diffusion models~\cite{park2025facial,he2024diff} and geometric priors~\cite{yang2024g,cao2023achieving} suggest promising directions, yet the field lacks frameworks characterizing achievable privacy-utility frontiers. Future priorities include: (a) standardized evaluation protocols; (b) information-theoretic analysis of privacy-utility trade-offs; and (c) video extensions with temporal consistency guarantees via 3D-aware or NeRF approaches~\cite{wen2023divide,cao2023achieving}.

\begin{table}[t]
\centering
\caption{\change{\textbf{Overview of primary face De-ID evaluation datasets}, grouped by modality. Env.: C = Controlled, S = Semi-controlled, W = In-the-wild.}}
\vspace{-1mm}
\setlength{\tabcolsep}{3pt}
\renewcommand{\arraystretch}{1.15}
\resizebox{1.0\linewidth}{!}{
\begin{tabular}{lccccl}
\toprule
\textbf{Dataset} & \textbf{\#Img/Vid} & \textbf{\#IDs} & \textbf{Resolution} & \textbf{Env.} & \textbf{Annotations} \\
\midrule
\multicolumn{6}{l}{\textit{\textbf{RGB Images}}} \\
\midrule
LFW~\cite{huang2008labeled}            & 13{,}233       & 5{,}749   & 250$\times$250   & W & identity \\
CelebA~\cite{liu2015deep}              & 202{,}599      & 10{,}177  & 178$\times$218   & W & identity, 40 attr., 5 lmk \\
CelebA-HQ~\cite{karras2018progressive} & 30{,}000       & ---       & 1024$\times$1024 & W & 40 attr., 5 lmk \\
FFHQ~\cite{karras2019style}            & 70{,}000       & ---       & 1024$\times$1024 & W & --- \\
VGGFace2~\cite{cao2018vggface2}        & 3.31M          & 9{,}131   & varies           & W & identity, pose, age \\
CASIA-WebFace~\cite{yi2014learning}    & 494{,}414      & 10{,}575  & varies           & W & identity \\
MS-Celeb-1M~\cite{guo2016ms}           & 10M            & 100{,}000 & varies           & W & identity \\
MegaFace~\cite{kemelmacher2016megaface}& 1M             & 690{,}572 & varies           & W & identity \\
PubFig~\cite{kumar2009attribute}       & 58{,}797       & 200       & varies           & W & 73 attr. \\
AgeDB~\cite{moschoglou2017agedb}       & 16{,}488       & 568       & varies           & W & identity, age, gender \\
CFP~\cite{sengupta2016frontal}         & 7{,}000        & 500       & varies           & W & frontal/profile \\
FERET~\cite{phillips1998feret}         & 14{,}126       & 1{,}199   & 256$\times$384   & C & pose, expre. \\
Multi-PIE~\cite{4813399}               & 750{,}000      & 337       & varies           & C & 15 poses, 19 illu., 6 expre. \\
MORPH~\cite{ricanek2006morph}          & 55{,}134       & 13{,}000  & varies           & S & age, gender, race \\
RaFD~\cite{langner2010presentation}    & 8{,}040        & 67        & 681$\times$1024  & C & 8 expre., gaze, 5 poses \\
CK+~\cite{lucey2010extended}           & 593 seq.       & 123       & 640$\times$490   & C & expre., AU \\
AffectNet~\cite{mollahosseini2017affectnet} & 450{,}000 & ---       & varies           & W & valence-arousal, 8 expre. \\
LADN~\cite{gu2019ladn}                 & 635            & ---       & varies           & W & makeup style \\
\midrule
\multicolumn{6}{l}{\textit{\textbf{RGB Video}}} \\
\midrule
YTF~\cite{wolf2011face}                & 3{,}425 vid.   & 1{,}595   & varies           & W & identity \\
VoxCeleb~\cite{nagraniy2017voxceleb} & 150K vid. & 6{,}112 & varies & W & speaker, audio \\
VidTIMIT~\cite{sanderson2009multi}     & 430 vid.       & 43        & 512$\times$384   & C & speaker, audio \\
Ego4D~\cite{grauman2022ego4d}          & 3{,}670 hrs    & ---       & varies           & W & narration, activity \\
\midrule
\multicolumn{6}{l}{\textit{\textbf{3D Face}}} \\
\midrule
Bosphorus~\cite{savran2008bosphorus}   & 4{,}666 scans  & 105       & 3D mesh          & C & 35 expre., AU, occlusion \\
BU-3DFE~\cite{yin20063d}               & 2{,}500 scans  & 100       & 3D mesh          & C & 6 expre. $\times$ 4 levels \\
Eurecom Kinect~\cite{min2014kinectfacedb} & 936        & 52        & RGB-D            & C & pose, expre., occlusion \\
SIAT-3DFE~\cite{ye2020siat}            & 8{,}000 scans  & 500       & 3D mesh          & C & expre. \\
\midrule
\multicolumn{6}{l}{\textit{\textbf{Physiological (rPPG / Affective)}}} \\
\midrule
PURE~\cite{stricker2014non}            & 60 vid.        & 10        & 640$\times$480   & C & heart rate ground truth \\
OBF~\cite{li2018obf}                   & 200 vid.       & 100       & varies           & C & heart rate, atrial fibrillation \\
AVEC~\cite{valstar2013avec,valstar2014avec} & ---       & 84+       & varies           & S & depression score \\
\bottomrule
\end{tabular}
}
\label{tab:datasets}
\vspace{-4mm}
\end{table}

\section{Evaluation Protocols}
\label{sec:protocol}

Face De-ID lacks the standardized benchmarks and metrics characteristic of mature vision tasks. Instead, evaluation practice has evolved heterogeneously, shaped by diverse threat models and application contexts across the three domains. TABLE~\ref{tab:methods_summary} summarizes the evaluation protocols of the 112 surveyed methods, revealing both remarkable breadth and concerning fragmentation. We first review the underlying datasets, then analyze the metrics used to assess privacy, utility, and visual quality.

\begin{table*}
\centering
\caption{\textbf{Overview of the face De-ID landscape highlighting the heterogeneity in evaluation protocols}. Methods are organized by domain (Physical, Sensor, Digital) and publication year, summarizing the specific datasets, performance metrics (Privacy$^a$, Utility$^b$, and Quality$^c$), and key innovations.}
\renewcommand{\arraystretch}{1.1}
\setlength{\tabcolsep}{4pt}
\resizebox{1.0\linewidth}{!}{
\begin{tabular}{lclcccl}
\toprule
\textbf{Method} & \textbf{Year} & \textbf{Dataset} & \textbf{Privacy}$^a$ & \textbf{Utility}$^b$ & \textbf{Quality}$^c$ & \textbf{Key Innovation} \\
\hline
\multicolumn{7}{c}{\textit{\textbf{Physical-Domain Face De-ID: Real-world Adversarial Modifications}} (Sec.~\ref{sec:physical})} \\
\hline
AdvEyeglass~\cite{sharif2016accessorize} & 2016 & PubFig~\cite{kumar2009attribute} & Success Rate ($\uparrow$): 31.0\% & \xmark & \xmark & Eyeglass-frame perturbation\\

AGNs~\cite{sharif2019general} & 2019 & PubFig~\cite{kumar2009attribute} & Success Rate ($\uparrow$): 70.0\%  & \xmark & \xmark & GAN-based natural eyeglasses \\

OAP~\cite{pautov2019adversarial} & 2019 & CASIA~\cite{yi2014learning} & Similarity ($\uparrow$) & \xmark & \xmark & Grayscale patches for ArcFace \\

VLA~\cite{shen2019vla} & 2019 &  CusFace, LFW~\cite{huang2008labeled} & Success Rate ($\uparrow$): 85.6\%  & \xmark & Similarity, Distance, SelfDis & Visible light-based physical attack \\

ALPA~\cite{nguyen2020adversarial} & 2020 & —- & Success Rate ($\uparrow$): 92.0\% & \xmark & \xmark & Adversarial light projection\\ 

AdvHat~\cite{komkov2021advhat} & 2021 & CASIA~\cite{yi2014learning} & Similarity ($\uparrow$) & \xmark & \xmark &  Hat-based surface attack \\

AdvMakeup~\cite{yin2021adv} & 2021 & LFW~\cite{huang2008labeled}, LADN~\cite{gu2019ladn} &  Success Rate ($\uparrow$): 63.74\% & \xmark & \xmark & Physical eye-shadow makeup attack\\

FaceAdv~\cite{shen2021effective} & 2021 & LFW~\cite{huang2008labeled}, VolFace~\cite{shen2021effective} & Success Rate ($\uparrow$): 100\% & \xmark & \xmark & Multi-shape sticker GAN  \\

GenAP~\cite{xiao2021improving} & 2021 &  LFW~\cite{huang2008labeled}, CelebA-HQ~\cite{karras2018progressive} & Success Rate ($\uparrow$): $\leq$99\%  & \xmark & \xmark & GAN manifold regularization \\

AdvSticker~\cite{wei2022adversarial} & 2022 &  LFW~\cite{huang2008labeled}, CelebA~\cite{liu2015deep} & Fooling Rate ($\uparrow$) & \xmark & \xmark & Attack using real-life stickers \\

Physical AX~\cite{singh2022powerful} & 2022 &  VGGFace2~\cite{cao2018vggface2} & Success Rate ($\uparrow$): $\leq$83\% & \xmark & \xmark & Threshold-based smoothness loss \\

AdvMask~\cite{zolfi2022adversarial} & 2022 & \makecell[l]{CASIA~\cite{yi2014learning}, CelebA~\cite{liu2015deep}, \\MS-Celeb-1M~\cite{guo2016ms} } & Recognition Rate ($\downarrow$) & \xmark & \xmark & Universal fabric mask patterns \\

PPAttack~\cite{wei2023simultaneously} & 2023 & LFW~\cite{huang2008labeled}, CelebA~\cite{liu2015deep} & Fooling Rate ($\uparrow$) & \xmark & \xmark & Simultaneously optimization \\

AT3D~\cite{yang2023towards} & 2023 & LFW~\cite{huang2008labeled}, CelebA-HQ~\cite{karras2018progressive} & Success Rate ($\uparrow$): $\leq$100\% & \xmark & \xmark &  Low-dim 3-DMM coefficient\\

Optical De-ID~\cite{li2023physical} & 2023 &  \makecell[l]{Bosphorus~\cite{savran2008bosphorus}, Eurecom~\cite{min2014kinectfacedb},\\ SIAT-3DFE~\cite{ye2020siat} } & Success Rate ($\uparrow$): $\leq$99\%  &  \xmark &  RMSE & Adversarial-illumination 3D attack \\

PadvFace~\cite{zheng2023robust} & 2023 & LFW~\cite{huang2008labeled} & Success Rate ($\uparrow$): 31\% & \xmark & \xmark & Curriculum-optimized sticker \\

SASMask~\cite{gong2023stealthy} & 2024 & \makecell[l]{LFW~\cite{huang2008labeled}, VGGFace2~\cite{cao2018vggface2}, \\ AgeDB~\cite{moschoglou2017agedb}, CFP~\cite{sengupta2016frontal} } & Success Rate ($\uparrow$): 43.4\% & \xmark & SSIM & Adversarial style optimization \\

Agile~\cite{wang2024invisible} & 2024 & LFW~\cite{huang2008labeled}, YTF~\cite{wolf2011face} & Success Rate ($\uparrow$): 80\% & \xmark & \xmark &  Invisible infrared laser perturbations \\

EAP~\cite{liu2024eap} & 2024 & LFW~\cite{huang2008labeled}, CelebA-HQ~\cite{karras2018progressive}  & Success Rate ($\uparrow$): 91.2\% & \xmark & \xmark & Meta-ensemble gradient extraction \\

AdvBandage~\cite{bhilare2024fooling} & 2024 & LFW~\cite{huang2008labeled} & Accuracy ($\downarrow$) & \xmark & \xmark & Physical attack using bandages   \\

ARA~\cite{zhang2024adversarial} & 2024 & VGGFace2~\cite{cao2018vggface2}, CelebA~\cite{liu2015deep} &  Success Rate ($\uparrow$): $\leq$99\% & \xmark & BRISQUE, NIQE & Physical natural adversarial lighting\\

Face3DAdv~\cite{yang2025face3dadv} & 2025 &  LFW~\cite{huang2008labeled}, CelebA-HQ~\cite{karras2018progressive} & Success Rate ($\uparrow$): $\leq$99\% &  \xmark & \xmark & Robust 3D adversarial patches \\

ProjAttacker~\cite{liu2025projattacker} & 2025 & LFW~\cite{huang2008labeled}, CelebA-HQ~\cite{karras2018progressive} & Success Rate ($\uparrow$): $\leq$98\% & \xmark & \xmark &  Configurable physical attack \\

\hline
\multicolumn{7}{c}{\textit{\textbf{Sensor-Domain Face De-ID: Acquisition-Level Privacy Protection}} (Sec.~\ref{sec:sensor})} \\
\hline
DefocusOptics~\cite{pittaluga2015privacy} & 2015 & —- & $k$-anonymity & Depth, Tracking & \xmark & Optical PSF-based k-anonymity \\

DefocusOptics$^+$~\cite{pittaluga2016pre} & 2016 & FERET~\cite{phillips1998feret} & $k$-anonymity  & Depth, Counting, Detection & \xmark & Pre-capture privacy-preserving optics\\

ISR~\cite{ryoo2017privacy} & 2017 & \makecell[l]{HMDB~\cite{kuehne2011hmdb}, DogCentric~\cite{iwashita2014first}, \\JPL-Interaction~\cite{ryoo2013first} } & Subjective & Action &  \xmark & Extreme low resolution \\

LRPrivacy~\cite{ryoo2018extreme} & 2018 & HMDB~\cite{kuehne2011hmdb}, DogCentric~\cite{iwashita2014first} & Subjective & Action & \xmark & Multi-view embedding \\

PrivHPE~\cite{hinojosa2021learning} & 2021 & \makecell[l]{LFW~\cite{huang2008labeled}, MS-Celeb-1M~\cite{guo2016ms}, \\ AgeDB~\cite{moschoglou2017agedb}, CFP~\cite{sengupta2016frontal}} & AUC ($\downarrow$) & Pose  & PSNR, SSIM & End-to-end optimize optical encoder\\

PrivHAR~\cite{hinojosa2022privhar} & 2022 & \makecell[l]{HMDB~\cite{kuehne2011hmdb}, VISPR~\cite{orekondy2017towards}, \\PA-HMDB~\cite{wu2020privacy} } & C-MAP ($\downarrow$), AUC ($\downarrow$) &  Action  & \xmark & Adversarial optical encoding \\

PrivPDE~\cite{tasneem2022learning} & 2022 & VGGFace2~\cite{cao2018vggface2}, NYUv2~\cite{silberman2012indoor} & AUC ($\downarrow$) & Depth, Action  & \xmark & Optimize phase mask \\

OpticalDR~\cite{pan2024opticaldr} & 2024 & \makecell[l]{CelebA~\cite{liu2015deep}, CK+~\cite{lucey2010extended}, \\ AVEC~\cite{valstar2013avec,valstar2014avec}} & AUC ($\downarrow$) & Emotion, Depression & \xmark & Depression signal preservation \\

DyPP~\cite{cheng2024learning} & 2024 & \makecell[l]{PubFig~\cite{kumar2009attribute}, LFW~\cite{huang2008labeled}, AgeDB~\cite{moschoglou2017agedb} } & Accuracy ($\downarrow$), AUROC ($\downarrow$) &   Counting, Pose, Detection & PSNR, SSIM & Time-varying optical privacy \\

PrivacyOptics~\cite{lopez2024privacy} & 2024 & CelebA-HQ~\cite{karras2018progressive}, FFHQ~\cite{karras2019style} & $L_{2}$ Distance ($\uparrow$) &  Attribute & FID & De-ID via learned optical encoder\\

\hline
\multicolumn{7}{c}{\textit{\textbf{Digital-Domain Face De-ID: Post-Capture Processing Methods}} (Sec.~\ref{sec:digital})} \\
\hline
\multicolumn{7}{l}{\textit{\changeminor{Handcrafted Distortion \& Filtering}} (Sec.~\ref{sec:naive})} \\
\hline
EFVAP~\cite{hudson1996techniques} & 1996 & --  & Subjective &  \xmark & \xmark & Gaussian blur and pixelation \\
De-ID Filter~\cite{boyle2000effects} & 2000 &  -- &  Protection Rate ($\uparrow$) & \xmark & \xmark &   Evaluation of filters\\
PUI~\cite{crowley2000things} & 2000 & --  & Subjective & \xmark &  \xmark & Eigen-space filtering\\

BFPP~\cite{neustaedter2006blur} & 2006 & -- & Awareness Rate ($\downarrow$) & \xmark &  \xmark & Home telepresence study\\

EmotionPreserve~\cite{letournel2015face} & 2015 & LFW~\cite{huang2008labeled} & Verification Rate ($\downarrow$)  & Expression, Gaze & \xmark & Variational adaptive filtering\\

\hline
\multicolumn{7}{l}{\textit{k-Anonymity \& Statistical Methods} (Sec.~\ref{sec:ksame})} \\
\hline
$k$-Same~\cite{newton2005preserving} & 2005 & FERET~\cite{phillips1998feret} & $k$-anonymity & \xmark & \xmark & First formal framework\\

$k$-Same-Select~\cite{gross2005integrating} & 2005 & FERET~\cite{phillips1998feret} &  Recognition Rate ($\downarrow$) &  Gender, Expression & \xmark & Incorporate utility constraints\\

$k$-Same-M~\cite{gross2006model} & 2006 & Multi-PIE~\cite{4813399}, FERET~\cite{phillips1998feret} & Recognition Rate ($\downarrow$) & Expression & \xmark & Model-based face De-ID\\

$\epsilon$-map DeID~\cite{gross2007towards} & 2007 & -- & Rank-1 Accuracy ($\downarrow$) & Gender, Expression & \xmark & k-anonymity guarantees\\

Multi-Factor DeID~\cite{gross2008semi} & 2008 & IMM Face Dataset~\cite{stegmann2003fame} & Recognition Rate ($\downarrow$) & Expression & \xmark & Multi-factor separation \\

$k$-Same-furthest~\cite{meng2014face} & 2014 & IMM Face Dataset~\cite{stegmann2003fame} & Recognition Rate ($\downarrow$) & Gender, Age, Expression & \xmark & Cluster furthest away\\

GARP-Face~\cite{du2014garp} & 2014 & MORPH Dataset~\cite{ricanek2006morph} & Recognition Accuracy ($\downarrow$) &  Race, Gender, Age & \xmark & Structured utility hierarchy \\

DiffPose-DeID~\cite{samarzija2014approach} & 2014 & IMM Face Dataset~\cite{stegmann2003fame} & Recognition Rate ($\downarrow$) & Pose, Emotion & \xmark & De-ID across different poses\\ 

Photorealistic-DeID~\cite{mosaddegh2014photorealistic} & 2014 & \makecell[l]{Multi-PIE~\cite{4813399}, MUCT~\cite{milborrow2010muct}, \\ PUT~\cite{kasinski2008put} } & ROC Analysis & Gender, Expression, Pose & PSNR & Component-based face synthesis \\

APFD~\cite{jourabloo2015attribute} & 2015 & \makecell[l]{FaceTracer~\cite{kumar2008facetracer}, FaceScrub~\cite{ng2014data}, \\ ND1~\cite{chang2003face}, FERET~\cite{phillips1998feret}, \\ CAS-PEAL~\cite{gao2007cas}, BioID~\cite{jesorsky2001robust}}   & Recognition Rate ($\downarrow$) & Attribute & PSNR  & Optimal weighted fusion \\

FIP-DeID~\cite{chi2015face} & 2015 & Multi-PIE~\cite{4813399} & Recognition Rate ($\downarrow$) & Gender, Race, Age, Expression & \xmark & Deep learning-based method\\

$k$-Same-Net~\cite{meden2018k} & 2018 & RaFD~\cite{langner2010presentation}, XM2VTS~\cite{messer1999xm2vtsdb}, CK+~\cite{lucey2010extended} & Recognition Rate ($\downarrow$) & Expression & \xmark & Generative realization of $k$-anonymity \\

ELEGANT-DeID~\cite{yan2019attributes} & 2019 & CelebA~\cite{liu2015deep} & $k$-anonymity & Age, Gender, Expression & \xmark & ELEGANT-based attribute transfer  \\

\hline
\multicolumn{7}{l}{\textit{Adversarial Perturbation Methods} (Sec.~\ref{sec:adversarial})} \\
\hline
EvolutionaryAttack~\cite{dong2019efficient} & 2019 &  LFW~\cite{huang2008labeled}, MegaFace~\cite{kemelmacher2016megaface} & MSE  ($\uparrow$) & \xmark & \xmark & Decision-based attack on recognition\\

P-FGVM~\cite{chatzikyriakidis2019adversarial} & 2019 & CelebA~\cite{liu2015deep} & Misclassification Rate ($\uparrow$) & \xmark &  MSSIM & Realism-penalized gradient \\

Makeup De-ID~\cite{zhu2019generating} & 2019 & Custom Dataset & Success Rate ($\uparrow$) & \xmark & \xmark & Makeup-based adversarial attack\\

AdvFaces~\cite{deb2020advfaces} & 2020 & CASIA~\cite{yi2014learning}, LFW~\cite{huang2008labeled} & Success Rate ($\uparrow$): 99.67\% &  \xmark & SSIM & GAN-based face synthesis \\

DFANet~\cite{zhong2020towards} & 2020 & \makecell[l]{LFW~\cite{huang2008labeled}, MS-Celeb-1M~\cite{guo2016ms}, \\ VGGFace2~\cite{cao2018vggface2},  CASIA~\cite{yi2014learning}, \\ IMDb-Face~\cite{wang2018devil} } & Success Rate ($\uparrow$) & \xmark & SSIM  & Dropout-based method \\

UAXs~\cite{amada2021universal} & 2021 & LFW~\cite{huang2008labeled}, VGGFace2~\cite{cao2018vggface2} & FMR ($\uparrow$)  &  \xmark &  \xmark &  Universal multi-identity spoofing\\

TIP-IM~\cite{yang2021towards} & 2021 & \makecell[l]{LFW~\cite{huang2008labeled}, MS-Celeb-1M~\cite{guo2016ms}, \\ MegaFace~\cite{kemelmacher2016megaface}} & Rank-N-T ($\downarrow$)  & \xmark & PSNR, SSIM, MMD & Principled distribution matching\\

Adv-Attribute~\cite{NEURIPS2022_dccbeb7a} & 2022 & FFHQ~\cite{karras2019style}, CelebA-HQ~\cite{karras2018progressive} & Success Rate ($\uparrow$) & \xmark &  FID, MSE & Compositional attribute editing \\

AMT-GAN~\cite{hu2022protecting} & 2022 & \makecell[l]{MT~\cite{li2018beautygan}, CelebA-HQ~\cite{karras2018progressive}, \\ LADN~\cite{gu2019ladn} } & Success Rate ($\uparrow$): 52.8\% & \xmark & FID, PSNR, SSIM & Makeup transfer (cycle-consistent)\\

FSAP~\cite{xue2023face} & 2023 & FFHQ~\cite{karras2019style}, CelebA~\cite{liu2015deep} & SPR ($\uparrow$) & Detection & MSE & StyleGAN W-space perturbation \\

CLIP2Protect~\cite{shamshad2023clip2protect} & 2023 & \makecell[l]{CelebA-HQ~\cite{karras2018progressive}, LADN~\cite{gu2019ladn}, \\ LFW~\cite{huang2008labeled} } &  Success Rate ($\uparrow$) &  \xmark   &  FID, PSNR, SSIM  &  Text-guided face De-ID \\

DiffProtect~\cite{liu2023diffprotect} & 2023 &  CelebA-HQ~\cite{karras2018progressive}, FFHQ~\cite{karras2019style} & Success Rate ($\uparrow$) & \xmark & FID & Semantic code optimization \\

Sibling-Attack~\cite{li2023sibling} & 2023 & CelebA-HQ~\cite{karras2018progressive}, LFW~\cite{huang2008labeled} & Success Rate ($\uparrow$) & \xmark & SSIM, MSE & Multi-task transferable attack    \\

SMAP~\cite{ma2023transferable} & 2023 & \makecell[l]{CelebA-HQ~\cite{karras2018progressive}, LFW~\cite{huang2008labeled}, \\VGGFace2~\cite{cao2018vggface2} } & Success Rate ($\uparrow$) & \xmark & FID, LPIPS, SSIM & Spatial mutable adversarial patch  \\

GMAA~\cite{li2023discrete} & 2023 & CelebA-HQ~\cite{karras2018progressive}, LFW~\cite{huang2008labeled} & Success Rate ($\uparrow$) & \xmark &  \xmark & Generalized manifold (AU-aware) \\

Chameleon~\cite{chow2024personalized} & 2024 & FaceScrub~\cite{ng2014data}, LFW~\cite{huang2008labeled} & Success Rate ($\uparrow$) & Pose, Expression, Age, Gender & FID, LPIPS, SSIM, User Study & Universal privacy protection mask\\

ADA~\cite{hu2024toward} & 2024 & CelebA-HQ~\cite{karras2018progressive}, LFW~\cite{huang2008labeled} & Success Rate ($\uparrow$) & \xmark & SSIM, PSNR, FID & Injection and denoising \\

FE-DeID~\cite{hanawa2024face} & 2024 & \makecell[l]{LFW~\cite{huang2008labeled}, CelebA~\cite{liu2015deep}, \\ GFSG2~\cite{karras2020analyzing}} & Success Rate ($\uparrow$) & \xmark & PSNR, SSIM, LPIPS & Feature embedding via CNN  \\

WeakenDiff~\cite{salar2025enhancing} & 2025 & CelebA-HQ~\cite{karras2018progressive}, LADN~\cite{gu2019ladn} & Success Rate ($\uparrow$): 79.2\% & \xmark & FID, PSNR, SSIM & Counter-purification targeting \\

DPA~\cite{zhou2025improving} & 2025 & CelebA-HQ~\cite{karras2018progressive}, LFW~\cite{huang2008labeled} & Success Rate ($\uparrow$): $\leq$98.2\%   & \xmark & \xmark &  Diverse parameters augmentation \\

Adv-Inversion~\cite{wang2025adv} & 2025 & FFHQ~\cite{karras2019style}, CelebA-HQ~\cite{karras2018progressive} & Success Rate ($\uparrow$), Rank-N-T ($\downarrow$) & \xmark & FID, SSIM, PSNR & GAN inversion framework \\

Adv-CPG~\cite{wang2025advcpg} & 2025 & \makecell[l]{FGID~\cite{huang2024consistentid}, FFHQ~\cite{karras2019style},\\ CelebA-HQ~\cite{karras2018progressive}, LFW~\cite{huang2008labeled}} & Success Rate ($\uparrow$) & \xmark & FID, SSIM, PSNR & Double-layer encryption \\

\hline
\multicolumn{7}{l}{\textit{Generative Model-Based Methods} (Sec.~\ref{sec:gennerative})} \\
\hline
Generative-DeID~\cite{brkic2017know} & 2017 & CCP~\cite{yang2014clothing}, Human3.6M~\cite{ionescu2013human3} & Accuracy ($\downarrow$)  & Garment Shape, Silhouette & Naturalness Score & GAN-based full-body De-ID \\

PP-GAN~\cite{wu2019privacy} & 2019 &  MORPH Dataset~\cite{ricanek2006morph} & De-ID Rate ($\uparrow$) & Detection, Age & SSIM  & Verificator+regulator paradigm \\

\hline
\multicolumn{7}{r}{\textit{Continued on next page}} \\
\end{tabular}}
\label{tab:methods_summary}
\end{table*}

\begin{table*}[t]
\renewcommand{\arraystretch}{1.1}
\setlength{\tabcolsep}{4pt}
\resizebox{1.0\linewidth}{!}{
\begin{tabular}{lclcccl}
\hline
FATM~\cite{li2019identification} & 2019 & \makecell[l]{LFW~\cite{huang2008labeled}, PIPA~\cite{zhang2015beyond}, \\ VidTIMIT~\cite{sanderson2009multi}} & Accuracy Drop ($\uparrow$) & Expression & SSIM & Encoder-decoder architecture \\

DeepPrivacy~\cite{hukkelaas2019deepprivacy} & 2019 & FDF, WIDER FACE~\cite{yang2016wider} &  Subjective & Detection & FID & Conditional GAN with U-Net\\

AnonymousNet~\cite{li2019anonymousnet} & 2019 & CelebA~\cite{liu2015deep} & $k$-anonymity & Attribute & PSNR, SSIM, MS-SSIM & StarGAN + formal privacy \\

Live-DeID~\cite{gafni2019live} & 2019 & \makecell[l]{LFW~\cite{huang2008labeled}, CelebA~\cite{liu2015deep}, \\ PubFig~\cite{kumar2009attribute}, CelebA-HQ~\cite{karras2018progressive}} & Accuracy ($\downarrow$) & Attribute & SSIM &  First video De-ID \\

CIAGAN~\cite{maximov2020ciagan} & 2020 & \makecell[l]{CelebA~\cite{liu2015deep}, MOTS~\cite{voigtlaender2019mots}, \\ LFW~\cite{huang2008labeled}} & Recall ($\uparrow$) & Detection & FID & Conditional GAN-Based De-ID \\

EPD-Net~\cite{aggarwal2020epd} & 2020 &  RaFD~\cite{langner2010presentation} &  De-ID Rate ($\uparrow$) & Emotion & SSIM & pix2pix-based face De-ID\\

Thermal-F De-ID~\cite{lin2020identification} & 2020 & Custom dataset & Accuracy ($\downarrow$) & Detection & \xmark & Novel feature extraction \\

FaceSwap~\cite{yang2021systematical} & 2021 & FFHQ~\cite{karras2019style} & ID Similarity ($\downarrow$)  & Detection, Expression & FID, LPIPS & Attribute disentanglement\\

PIDIM~\cite{cao2021personalized} & 2021 &  \makecell[l]{CelebA-HQ~\cite{karras2018progressive}, FFHQ~\cite{karras2019style}, \\ CASIA~\cite{yi2014learning}} &  De-ID Rate ($\uparrow$) & Detection &  \xmark & Password-based reversibility \\

DeID-GAN~\cite{agarwal2021privacy} & 2021 & \makecell[l]{RaFD~\cite{langner2010presentation}, XM2VTS~\cite{messer1999xm2vtsdb}, \\ CelebA~\cite{liu2015deep}, FFHQ~\cite{karras2019style}} & DMOS ($\uparrow$)  & Emotion, Attribute & FMOS & Three-stage pipeline with StyleGAN\\

AutoDeID~\cite{kuang2021effective} & 2021 & \makecell[l]{LFW~\cite{huang2008labeled}, VGGFace2~\cite{cao2018vggface2},\\ CelebA-HQ~\cite{karras2018progressive}} & Re-ID Recall ($\uparrow$) & Detection, Attribute  & FID & Combining k-anonymity with GAN\\

SF-GAN~\cite{li2021sfgan} & 2021 & CelebA-HQ~\cite{karras2018progressive}, FFHQ~\cite{karras2019style} & SSIM ($\downarrow$) & Expression, Gender, Hairstyle, Glasses & \xmark & Multi-attribute preservation \\

A$^3$GAN~\cite{zhai2022a3gan} & 2022 & \makecell[l]{CelebA~\cite{liu2015deep}, WIDER FACE~\cite{yang2016wider}, \\ ExpW~\cite{zhang2018facial}, NTHU-DDD~\cite{weng2016driver}} & Accuracy ($\downarrow$) & Detection, Expression & FID & Attribute editing-based De-ID \\

IdentityMask~\cite{wen2022IdentityMask} & 2022 &  VoxCeleb~\cite{nagraniy2017voxceleb} & CSIM ($\downarrow$) & Pose, Expression, Landmark & FID & Reversible face video De-ID \\

UU-Net~\cite{proencca2022uu} & 2022 & \makecell[l]{YTF~\cite{wolf2011face}, P-DESTRE~\cite{kumar2020p},\\ BIODI, MARS~\cite{zheng2016mars}} & AUC ($\downarrow$)  & Pose, Lighting, Background, Expression & \xmark & ROI steganography (surveillance)  \\

MVC-DeID~\cite{cao2023achieving} & 2023 & -- & ID Distance ($\uparrow$) &  Detection  &  PSNR, SSIM, LPIPS  &  3D-aware generators  \\

CPP-DeID~\cite{meden2023face} & 2023 & \makecell[l]{CelebA-HQ~\cite{karras2018progressive}, RaFD~\cite{langner2010presentation}, \\XM2VTS~\cite{messer1999xm2vtsdb}, AffectNet~\cite{mollahosseini2017affectnet}} & AUC ($\downarrow$), De-ID score ($\uparrow$) & Gender, Expression & FID, LPIPS, SSIM, MSE & Controllable privacy protection \\

DartBlur~\cite{jiang2023dartblur} & 2023 & \makecell[l]{WIDER FACE~\cite{yang2016wider}, FDDB~\cite{jain2010fddb}, \\ CrowdHuman~\cite{shao2018crowdhuman}} & Subjective & Face Detection & PSNR, SSIM & Learnable U-Net-based blur model \\

DeID rPPG~\cite{savic2023identification} & 2023 & PURE~\cite{stricker2014non}, OBF~\cite{li2018obf} & Accuracy ($\downarrow$)  & rPPG signal & PSNR, SSIM & De-ID preserving rPPG signals \\

IDeudemon~\cite{wen2023divide} & 2023 &  CelebA-HQ~\cite{karras2018progressive}, FFHQ~\cite{karras2019style} & $L_{2}$ Distance ($\uparrow$) & Pose, Expression  & PSNR, SSIM, FID & Parametric NeRF De-ID\\

FALCO~\cite{barattin2023attribute} & 2023 & CelebA-HQ~\cite{karras2018progressive}, LFW~\cite{huang2008labeled} & Re-ID Rate ($\downarrow$) &  Detection & FID & Latent code optimization \\

Egocentric DeID~\cite{puangthamawathanakun2023towards} & 2023 & Ego4D~\cite{grauman2022ego4d} & Similarity ($\downarrow$), Accuracy ($\downarrow$) & \xmark & FID, LPIPS, SSIM & Consistent identity replacement\\

StyleGAN-DeID~\cite{khorzooghi2023examining} & 2023 & CelebA~\cite{liu2015deep}, LFW~\cite{huang2008labeled} & Success Rate ($\uparrow$) & Detection, Gender, Pose, Expression, Age &  \xmark & StyleGAN's style-mixing \\

Verifiable DeID~\cite{park2024verifiable} & 2024 & LFW~\cite{huang2008labeled}, FFHQ~\cite{karras2019style} & Selection Rate ($\downarrow$)  & \xmark &  FID, SSIM, BRISQUE & Verifiable proof framework\\

Disguise~\cite{cai2024disguise} & 2024 & \makecell[l]{VGGFace2~\cite{cao2018vggface2}, LFW~\cite{huang2008labeled}, \\ CelebA-HQ~\cite{karras2018progressive}} & TPR ($\downarrow$)  & Landmark, Gaze, Emotion & SER-FIQ &  DP-inspired pseudo-identities \\

RBGAN~\cite{zhang2024rbgan} & 2024 & \makecell[l]{CelebA-HQ~\cite{karras2018progressive}, FFHQ~\cite{karras2019style}, \\ CASIA~\cite{yi2014learning}, LFW~\cite{huang2008labeled} } &  CSIM ($\downarrow$): 23.0\% & Face Detection, Attribute & SSIM, PSNR, FID &  Symmetric-consistency guidance \\

DiffAM~\cite{sun2024diffam} & 2024 &  CelebA-HQ~\cite{karras2018progressive}, LADN~\cite{gu2019ladn} & Success Rate ($\uparrow$) & \xmark & FID, PSNR, SSIM & Diffusion makeup adversarial \\

FaceCaricature~\cite{laishram2024face} & 2024 & CelebA-HQ~\cite{karras2018progressive} & Recognition Rate ($\downarrow$) & Attribute & LPIPS, SSIM, FID & Face De-ID using face caricatures \\

Veil Privacy~\cite{pang2024veil} & 2024 & \makecell[l]{BioID~\cite{jesorsky2001robust}, ORL~\cite{samaria1994parameterisation}, \\ LFW~\cite{huang2008labeled}, CelebA~\cite{liu2015deep}} & SSIM ($\downarrow$), User Study & Classification & SSIM & Random pixel flipping\\

FaceMotionPreserve~\cite{zhu2024facemotionpreserve} & 2024 & PD Dataset & Similarity ($\downarrow$) & Medical Info & \xmark & GAN-based face swapping \\

Diff-Privacy~\cite{he2024diff} & 2024 &  CelebA-HQ~\cite{karras2018progressive}, LFW~\cite{huang2008labeled} & Re-ID Rate ($\downarrow$), mAP ($\downarrow$), Accuracy ($\downarrow$) & Pose, Expression, Age, Gender & FID, KID, SSIM, LPIPS & Unified diffusion framework \\

G$^2$Face~\cite{yang2024g} & 2024 & \makecell[l]{CelebA-HQ~\cite{karras2018progressive}, LFW~\cite{huang2008labeled}, \\ FFHQ~\cite{karras2019style} } &  CSIM ($\downarrow$), TAR ($\downarrow$) & Detection, 3D Shape, Expression, Pose & FID & Reversible face De-ID\\

DRGAN~\cite{liu20253d} & 2025 & \makecell[l]{BU-3DFE~\cite{yin20063d}, BU-4DFE~\cite{zhang2013high}, \\ Bosphorus~\cite{savran2008bosphorus}} &  ACC ($\downarrow$), EER ($\uparrow$)  & Biometric, Expression, Gender, Ethnicity & \xmark & 2D/3D face De-ID \\

Face-DeID-Net~\cite{zeng2025face} & 2025 & CelebA-HQ~\cite{karras2018progressive}, MyStyle~\cite{nitzan2022mystyle} & Re-ID Rate ($\downarrow$) & Attribute &  IQA & Identity extraction and removal \\

Synthetic DeID~\cite{park2025facial} & 2025 & CelebA-HQ~\cite{karras2018progressive}, FFHQ~\cite{karras2019style} & SID ($\uparrow$), TID ($\downarrow$) & Landmark, Emotion &  LPIPS & Training-free identity editing\\

VCA-DeID~\cite{kim2025visual} & 2025 & \makecell[l]{CelebAMask-HQ~\cite{lee2020maskgan}, FFHQ~\cite{karras2019style}, \\ LFW~\cite{huang2008labeled}} & IS ($\downarrow$), VR ($\downarrow$), EER ($\downarrow$) & Attribute, Gender, Race & FID, FIQ  &  Image-text aligned attributes \\

Swapping De-ID~\cite{kung2025face} & 2025 & CelebA-HQ~\cite{karras2018progressive}, FFHQ~\cite{karras2019style} & Re-ID Rate ($\downarrow$), Face Distance ($\uparrow$) & Pose, Gaze, Expression  & IQA & Face swapping for De-ID  \\
\bottomrule
\end{tabular}}
\label{tab:xxx}
\vspace{-5mm}
\end{table*}

\vspace{-4mm}
\subsection{Evaluation Datasets}
TABLE~\ref{tab:datasets} summarizes the principal characteristics of the most widely adopted face De-ID datasets, grouped by modality and including capture environment, scale, resolution, and annotation richness. Across the 112 surveyed methods, a clear concentration emerges: LFW~\cite{huang2008labeled}, CelebA / CelebA-HQ~\cite{liu2015deep,karras2018progressive}, VGGFace2~\cite{cao2018vggface2}, and FFHQ~\cite{karras2019style} together account for the majority of evaluations. 
Task-specific benchmarks address narrower utility goals, \emph{e.g.}, RaFD~\cite{langner2010presentation} and AffectNet~\cite{mollahosseini2017affectnet} for expression, Bosphorus~\cite{savran2008bosphorus} for 3D-aware methods, and PURE~\cite{stricker2014non} for rPPG-preserving De-ID, while a smaller subset evaluates privacy--utility trade-offs in activity recognition (HMDB~\cite{kuehne2011hmdb}, DogCentric~\cite{iwashita2014first}).

\changeminor{Beyond adoption statistics, these datasets exhibit structural properties that determine their suitability. \emph{(i) Web-scraped, celebrity-dominated composition}: the dominant benchmarks (LFW, CelebA(-HQ), VGGFace2, MS-Celeb-1M, MegaFace) contain adult, quasi-frontal, well-lit faces with demographic skew, so results on them overstate robustness for children, older adults, and under-represented groups. \emph{(ii) Identity as the only universal annotation}: utility labels (expression, AUs, gaze, physiological signals) exist only in separate, smaller, often lab-controlled corpora; this asymmetry is a root cause of the fragmentation in Fig.~\ref{fig:fragmentation}, since privacy scores are comparable across papers while utility must be evaluated on disjoint datasets. Resolution and labeling further partition roles: LFW supports recognition-based privacy evaluation but not generative-quality assessment. CelebA(-HQ) and FFHQ provide the resolution and attribute labels needed by generative methods, though FFHQ lacks identity labels and thus cannot support formal privacy measurement. FERET and Multi-PIE remain the testbeds for factor-isolating $k$-anonymity analyses. \emph{(iii) Static RGB stills as the default modality}: video, 3D, and physiological resources (YTF, Ego4D; Bosphorus, BU-3DFE; PURE, OBF) are scarce and small-scale, explaining why temporal consistency and micro-level utility signals remain under-evaluated.}

\changeminor{In total, 66 distinct datasets appear across the method-level protocols catalogued in TABLE~\ref{tab:methods_summary}, of which TABLE~\ref{tab:datasets} distills the 29 primary, recurrently adopted ones. Despite this apparent abundance, three critical gaps remain. First, \textbf{evaluation fragmentation across domains} (TABLE~\ref{tab:methods_summary}, Fig.~\ref{fig:fragmentation}): physical-domain methods (23 techniques) report privacy exclusively, sensor-domain methods (10 techniques) universally report both privacy and utility, and digital-domain methods are the most heterogeneous.}
Second, \textbf{lack of multi-modal semantic annotations}: as TABLE~\ref{tab:datasets} shows, no single dataset jointly provides identity, fine-grained expression, demographics, and physiological labels, forcing researchers to evaluate utility preservation across heterogeneous datasets (\emph{e.g.}, expression on RaFD, attributes on CelebA), which impedes fair benchmarking. Third, \textbf{paucity of child-centric resources}: although datasets such as FG-NET~\cite{panis2014overview} (1{,}002 images, 82 subjects), ITWCC~\cite{deb2020child} (1{,}705 images, 304 subjects), and CLF~\cite{deb2018longitudinal} exist, they lack the scale and diversity required for training and analyzing, leaving vulnerable populations underserved.

\vspace{-3mm}
\subsection{Evaluation Metrics}
Face De-ID evaluation balances three competing objectives: privacy protection, utility preservation, and visual quality. \change{Fig.~\ref{fig:fragmentation} summarizes the evaluation landscape: the left reports how often each aspect is measured per domain, and the right ranks the 69 distinct metrics extracted from existing papers by adoption. Two forms of fragmentation are evident: the set of reported aspects differs across domains, and the choice of metric within each aspect is largely method-specific. The metric families below are organized along the three axes; per-metric adoption is deferred to Fig.~\ref{fig:fragmentation}.}

\subsubsection{Privacy Protection Evaluation Metrics}
\begin{figure*}[t]
\centering
\includegraphics[width=\linewidth]{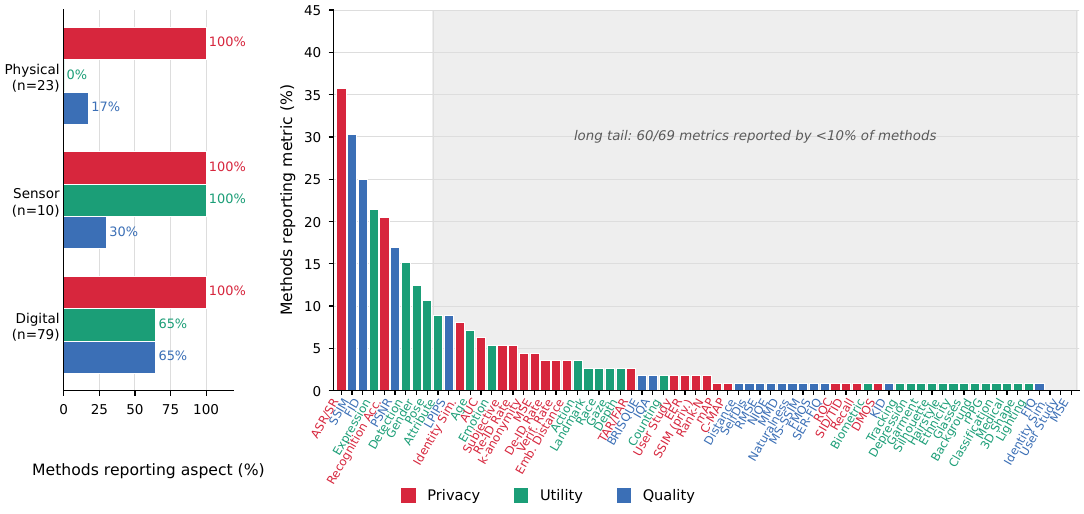}
\vspace{-6mm}
\caption{\textbf{Evaluation-protocol fragmentation across the surveyed face De-ID methods.}
\emph{Left}: per-domain reporting rates of the three evaluation aspects. Physical-domain methods ($n{=}23$) report privacy universally but utility never and quality rarely; sensor-domain methods report privacy and utility universally and quality in 30\% of cases; digital-domain methods report privacy universally but utility and quality in only 65\% of cases each.
\emph{Right}: adoption of the 69 distinct metrics used by existing papers, ranked by reporting share and colored by aspect. Only ASR ($\sim$36\%), SSIM ($\sim$30\%), and FID ($\sim$25\%) exceed a quarter of methods, while 60 metrics (grey) fall below 10\%.}
\vspace{-5mm}
\label{fig:fragmentation}
\end{figure*}

\textbf{Identity Verification and Recognition Metrics.} The most direct privacy assessment employs FR systems as adversaries. True Acceptance Rate (TAR) at fixed False Acceptance Rate (FAR) thresholds (commonly FAR=0.1\% or 0.01\%) measures genuine pair acceptance, with lower TAR indicating stronger De-ID. Equal Error Rate (EER), where FAR equals False Rejection Rate, provides a threshold-independent metric pushed toward 50\% (random chance) by effective De-ID. Rank-N accuracy assesses whether the true identity appears in the top-N retrievals. Attack Success Rate (ASR), or fooling rate, quantifies adversarial effectiveness: for dodging it measures the proportion of pairs failing verification, and for impersonation the rate of successful targeted misclassification. \change{As Fig.~\ref{fig:fragmentation} shows, ASR is the single most widely reported metric ($\sim$36\%) and dominates the physical domain.}

\noindent\textbf{Embedding Space Metrics.} Latent-space measurements provide fine-grained privacy quantification beyond binary verification. Cosine Similarity (CSIM) and $L_{2}$ Distance between original and De-ID embeddings measure feature-space proximity, with lower similarity and larger distance indicating stronger privacy. Source/Target Identity Distance (SID/TID) extends these to targeted scenarios, with SID maximized against the original identity and TID minimized toward the target. Maximum Mean Discrepancy (MMD) quantifies distributional divergence between embedding sets, enabling dataset-level assessment valuable for evaluating transferability.

\noindent\textbf{Formal Privacy Guarantees.} $k$-anonymity-based methods enforce group indistinguishability, upper-bounding recognition accuracy at $1/k$; De-ID rate measures the proportion of faces meeting this bound or failing verification at a given threshold. Near-perfect De-ID (TPR@FPR) reports the re-identification rate at extremely low False Positive Rates (e.g., FPR=0.02\%), relevant for high-security scenarios where even rare breaches are unacceptable.

\noindent\textbf{Human Perceptual Metrics.} Algorithmic metrics may diverge from human recognition. Human recognition rate through user studies provides ground-truth privacy assessment, and Mean Opinion Scores (MOS) capture subjective De-ID and naturalness. Such evaluations are critical for social acceptability (used by $\sim$15\% of surveyed methods).

\subsubsection{Utility Preservation Evaluation Metrics}
Utility metrics assess the retention of task-relevant, non-identity signals, and are coupled to the target application.

\noindent\textbf{Facial Attribute Preservation.} Attribute classifiers measure retention of soft biometrics, with classification accuracy for gender, age, and race quantifying utility degradation.
Landmark localization error (normalized mean error over 68 fiducial points) assesses geometric structure preservation, critical for gaze estimation and medical diagnostics. Detection metrics (mAP, recall) evaluate whether De-ID faces remain detectable, essential for surveillance utility.

\noindent\textbf{Expression and Affect Recognition.} Emotion-preserving methods report accuracy on expression datasets (e.g., CK+~\cite{lucey2010extended}, AVEC~\cite{valstar2013avec,valstar2014avec}), with strong methods retaining $\textgreater85\%$ of original performance. Action Unit (AU) detection~\cite{friesen1978facial} provides fine-grained assessment of facial movement preservation.

\noindent\textbf{Pose and Gaze Estimation.} Head pose error (mean angular error in yaw/pitch/roll) and gaze estimation error quantify geometric utility, the latter critical for attention analysis and driver monitoring. Depth estimation metrics (RMSE, absolute relative error) assess 3D utility, particularly for the sensor-domain methods in TABLE~\ref{tab:methods_summary} that prioritize depth.

\noindent\textbf{Remote Physiological Signal Preservation.}
Remote photoplethysmography (rPPG) enables contactless extraction of cardiovascular signals from facial videos. Evaluation metrics include pulse-waveform SNR, heart-rate MAE, and Pearson correlation of inter-beat interval sequences, with PURE~\cite{stricker2014non} and OBF~\cite{li2018obf} providing synchronized ground truth. DeID rPPG~\cite{savic2023identification} showed that conventional face De-ID severely degrades rPPG quality, motivating approaches that explicitly preserve frequency-domain characteristics.

\subsubsection{Visual Quality Metrics}
Visual quality metrics assess photorealism, naturalness, and artifact presence of the de-identified outputs.

\noindent\textbf{Pixel-Level Distortion Metrics.} Peak Signal-to-Noise Ratio (PSNR) measures pixel-wise fidelity but correlates poorly with perceptual quality for synthesized content. Structural Similarity Index (SSIM)~\cite{wang2004image} captures structural preservation (0-1 scale), and Multi-Scale SSIM extends it across spatial scales. \change{SSIM is the most widely reported quality metric ($\sim$30\%, Fig.~\ref{fig:fragmentation}), yet aligns poorly with human perception for GAN-synthesized or heavily perturbed images.}

\noindent\textbf{Perceptual / Distribution Metrics.} Fréchet Inception Distance (FID)~\cite{heusel2017gans} measures distributional divergence in Inception-v3~\cite{szegedy2016rethinking} feature space; Kernel Inception Distance (KID) is an unbiased alternative with better small-sample properties~\cite{he2024diff}; and LPIPS~\cite{zhang2018unreasonable} correlates strongly with human judgment. \change{FID ($\sim$25\%) is the third most reported metric overall; generative methods adopt FID and LPIPS almost universally, whereas adversarial methods rarely use them due to minimal pixel changes.}

\noindent\textbf{Image Quality Assessment.} No-reference metrics assess naturalness without a reference image. BRISQUE and NIQE detect artifact-indicative statistical anomalies~\cite{zhang2024adversarial,park2024verifiable}, with lower scores indicating higher naturalness, and Face Image Quality (FIQ) scores (e.g., SER-FIQ) evaluate recognizability for face recognition~\cite{cai2024disguise}, providing coupled privacy-utility insight.

\noindent\textbf{Human Perceptual Studies.} User studies provide ground-truth quality assessment where algorithmic proxies fail, via MOS for photorealism, the De-ID-specific FMOS and DMOS, and pairwise preference tests that reduce absolute-rating bias. \change{As Fig.~\ref{fig:fragmentation} confirms, such studies sit in the long tail of rarely adopted metrics despite their deployment importance, limited by cost and scalability.}

\vspace{-3mm}
\subsection{Performance Analysis}
Despite the evaluation fragmentation, cross-domain analysis of TABLE~\ref{tab:methods_summary} reveals discernible performance patterns.

Physical-domain methods achieve remarkably high attack success rates: 3D adversarial meshes (AT3D~\cite{yang2023towards}) and projection-based approaches (ProjAttacker~\cite{liu2025projattacker}) reach near-perfect rates (98--100\%) against FR models, while eyeglass-based methods (AdvEyeglass~\cite{sharif2016accessorize}, AGNs~\cite{sharif2019general}) trail at 31--70\% due to limited perturbation area. However, all 23 physical-domain methods report privacy metrics exclusively, reflecting their attack-centric origins.

Sensor-domain methods exhibit the most balanced privacy-utility trade-offs, with all 10 techniques reporting both privacy and utility. PrivHAR~\cite{hinojosa2022privhar} and PrivPDE~\cite{tasneem2022learning} reduce FR to near-random levels (AUC approaching 0.5) while keeping depth estimation within acceptable bounds, OpticalDR~\cite{pan2024opticaldr} preserves depression-related diagnostic signals, and DyPP~\cite{cheng2024learning} adds robustness to PSF inversion. This balance reflects the inherent advantage of optical encoding, which attenuates identity-discriminative frequencies while preserving task-relevant information.

Digital-domain methods exhibit the greatest heterogeneity across the $\sim$80 techniques catalogued. Handcrafted distortion and $k$-anonymity methods ($k$-Same~\cite{newton2005preserving}, $k$-Same-Select~\cite{gross2005integrating}) provide formal guarantees but limited utility and quality. Adversarial perturbation methods achieve high success rates but transfer poorly to black-box systems. Generative approaches deliver the best holistic performance: GAN-based (FALCO~\cite{barattin2023attribute}, CPP-DeID~\cite{meden2023face}, Disguise~\cite{cai2024disguise}) and diffusion-based methods (Diff-Privacy~\cite{he2024diff}, Synthetic DeID~\cite{park2025facial}) retain attribute preservation above 90\% while achieving effective De-ID and competitive quality, and hybrid adversarial-generative methods (DiffProtect~\cite{liu2023diffprotect}, WeakenDiff~\cite{salar2025enhancing}) point toward unified privacy-utility-quality optimization.

In summary, physical-domain methods achieve the strongest privacy but neglect utility; sensor-domain methods offer principled privacy-utility optimization through differentiable optical design; and digital-domain methods provide maximum flexibility, with generative approaches delivering the best holistic performance. The field shows a clear progression from ad-hoc distortion toward learned representations that jointly optimize competing objectives, yet the absence of standardized benchmarks remains a critical barrier to definitive cross-method comparison.

\section{Future Directions and Open Challenges}
	\label{sec:discussion}

While substantial progress has been achieved across the physical, sensor, and digital domains, critical challenges and emerging opportunities warrant focused investigation.

\vspace{1mm}
\noindent\textbf{Physical Domain: Bridging the Utility Gap.}
Sec.~\ref{sec:protocol} shows that physical-domain methods universally report privacy metrics without evaluating \textbf{utility preservation}, reflecting adversarial origins where recognition evasion supersedes downstream task enablement. Yet real-world deployment (\emph{e.g.}, healthcare monitoring, emotion-aware interaction) demands physical De-ID that preserves task-relevant attributes. Closing this gap requires two innovations: (1) multi-objective optimization that jointly maximizes privacy and minimizes utility degradation across tasks (expression recognition, gaze estimation, age perception) under manufacturability constraints (printability, material reflectance, durability) absent in digital methods; and (2) attribute-disentangled adversarial patterns that suppress identity-discriminative features while sparing utility-related regions. Social acceptability is a further concern: bandages and medical masks offer plausible cover, whereas adversarial eyeglasses and stickers may attract unwanted attention, undermining covert protection.

\vspace{1mm}
\noindent\textbf{Sensor Domain: Advancing Programmable Optics.}
Sensor-based face De-ID offers strong theoretical guarantees by preventing identity capture before digitization, yet current implementations rely on static or learned-but-fixed optics (Fig.~\ref{fig:sensor}), lacking \textbf{adaptive systems} that adjust privacy-utility trade-offs to scene context or task. Three directions warrant priority: (1) context-aware programmable optics that pair scene understanding (medical consultation versus public space) with real-time optical reconfiguration; (2) multi-modal sensor fusion combining privacy-encoded RGB with unencoded depth or infrared to preserve geometric utility (pose, tracking) while suppressing identity-rich texture; and (3) co-design of optical encoders and downstream networks through differentiable imaging pipelines. Programmable optics, however, incur latency, power, and cost penalties that limit deployment, so standardized protocols must assess privacy-utility trade-offs alongside computational efficiency, hardware complexity, and long-term robustness.

\vspace{1mm}
\change{\noindent\textbf{Digital Domain: Verifiable and Reversible Face De-ID.} 
Two coupled challenges define the next phase of digital-domain face De-ID: \emph{verifiability}, the ability to prove that a de-identified output meets a stated privacy criterion without revealing the original $\mathbf{I}$, and \emph{reversibility} (criterion~(4) of Sec.~\ref{sec:problem_formulation}), required by forensic, medical, and legal workflows but difficult to combine with strong irreversible guarantees. Existing reversible methods rely predominantly on password- or key-conditioned generative inversion~\cite{wen2022IdentityMask,yang2024g}, which is vulnerable to brute-force, key-leakage, and auxiliary-information attacks because the recovery mapping is learned rather than cryptographically protected. A promising direction is to integrate cryptographic primitives, particularly homomorphic encryption and secure multi-party computation, with generative models, so that the recovery oracle $\mathcal{R}(\cdot\,;k)$ is realized through provably secure protocols rather than learned mappings alone. Verifiability can be addressed in parallel through zero-knowledge proofs or audit logs that attest, without disclosing $\mathbf{I}$, that $\tilde{\mathbf{I}}$ was produced by a sanctioned $\mathcal{F}$ meeting a privacy threshold. }

\vspace{1mm}
\noindent\textbf{Addressing Micro-Level Utility Signals.}
Face De-ID research predominantly targets macro-level attributes (age, gender, expression), yet emerging healthcare, affective computing, and behavioral analysis applications demand preservation of \textbf{micro-level} signals far more sensitive to identity suppression. Three categories stand out. (1) \emph{Physiological signals}~\cite{savic2023identification} (rPPG, heart rate, respiration) require signal-aware methods with frequency-domain constraints or correlation losses validated across skin tones, lighting, and De-ID paradigms. (2) \emph{Micro-expressions} (fleeting movements revealing genuine emotion) require temporal robustness through optical-flow consistency and micro-expression-aware losses, supporting deception detection, mental health assessment, and human-robot interaction. (3) \emph{Neurological diagnostic features} (stroke asymmetries, Parkinsonian tremor, autism-related expression abnormalities) require interdisciplinary collaboration to extend the OpticalDR paradigm~\cite{pan2024opticaldr}, validate clinical accuracy, and earn medical community acceptance. Underlying all three is the fine-grained coupling between identity and utility, which calls for information-theoretic bounds on achievable privacy-utility trade-offs and principled, application-driven navigation of those bounds.

\vspace{1mm}
\noindent\textbf{Unified Evaluation Protocols and Benchmarks.}
The \textbf{evaluation fragmentation} identified in Sec.~\ref{sec:protocol} represents a primary bottleneck. Addressing these issues requires community-wide coordination: (1) curated multi-task datasets with comprehensive annotations (identity, age, gender, expression, landmarks); (2) standardized metrics spanning privacy (TAR@FAR, embedding distance), utility (attribute classification, landmark localization, expression recognition), and quality (FID, LPIPS, BRISQUE, NIQE); (3) reference implementations across all three domains with reproducible scripts; and (4) continuously updated leaderboards tracking SOTA performance. Beyond technical infrastructure, benchmark design must address fairness by stratifying performance across race, gender, and age, while adversarial robustness evaluation (JPEG compression, resizing, denoising) remains critical but underexplored. 

\vspace{1mm}
\noindent\textbf{Cross-Domain Co-Design for Privacy-First Sensing.}
The convergence of optical and algorithmic privacy mechanisms invites a rethinking of the sensing-to-analytics pipeline, with next-generation systems integrating the \textbf{physical-sensor-digital} stack to enforce privacy at the earliest stages of acquisition. Programmable sensors can implement De-ID in analog or early digital domains through neuromorphic architectures or event-based cameras that inherently suppress identity-rich spatial detail while retaining motion and scene dynamics. Differentiable imaging pipelines complement this by jointly optimizing optical elements (metasurfaces, diffractive optics), image signal processing, and downstream task networks under unified privacy-utility objectives. Realizing this vision requires differentiable models of the full image-formation process and optimization strategies that account for physical realizability, manufacturing tolerances, and deployment conditions, marking a shift from post-capture De-ID toward architectures where protection is intrinsic to the imaging system.

\vspace{1mm}
\noindent\textbf{Emerging Application Domains.}
Beyond traditional surveillance, access control, and social media, face De-ID must address \textbf{emerging domains} where conventional approaches fall short. Three stand out. (1) \emph{Autonomous vehicles and smart cities} deploy ubiquitous sensing and require perception systems that balance public safety against individual privacy. (2) \emph{Medical and telehealth} applications need domain-specific methods that preserve clinically relevant cues (expressions for neurological assessment, dermatological conditions for diagnosis) while preventing re-identification. (3) \emph{Metaverse and extended reality} introduce new attack surfaces, where avatars, behavioral patterns, and physiological signals (eye tracking, gestures, locomotion) enable identity inference beyond facial biometrics. Meeting these challenges demands collaboration across computer vision, cryptography, hardware design, and policy.

\noindent\textbf{Joint De-ID Across Face, Body, and Behavioral Channels.} Face De-ID alone is insufficient when other identity-bearing channels remain intact: body shape, gait, voice, eye gaze, and hand motions are each independently sufficient for re-identification, and recent surveys document active research on each~\cite{khan2024person,hanisch2025anonymization}. 
\changeminor{Face De-ID nonetheless retains substantial standalone value. The face is the most discriminative and most widely operationalized biometric cue~\cite{zhao2003face}, targeted by commercial recognition systems, watchlists, and web-scale search~\cite{laishram2025toward}, so its suppression removes the adversary with the greatest practical reach. In the data-release scenarios that most often motivate De-ID (photographs, medical and telehealth recordings~\cite{yang2022digital}, social-media imagery~\cite{liu2022privacy}), the face is often the only channel captured at recognition-grade quality; gait requires temporal sequences~\cite{sepas2022deep} and body-appearance matching degrades under clothing and viewpoint variation~\cite{zhang2015beyond}. Regulatory instruments such as the EU AI Act~\cite{EUAIAct2024} and GDPR~\cite{GDPR2016} further attach specific obligations to facial data. Face De-ID thus mitigates the dominant re-identification risk, while joint multi-channel anonymization remains necessary against adversaries with high-quality dynamic footage.}
Three open problems emerge at the intersection of face and these complementary channels: (1) \emph{joint anonymization frameworks} that suppress identity across face, body, and motion simultaneously without re-introducing leakage in any single channel; (2) \emph{temporal-consistency guarantees} for dynamic traits (\emph{e.g.}, gait, micro-expressions), where frame-wise face De-ID may still leak identity through motion signatures; and (3) \emph{cross-modal evaluation protocols} that quantify residual leakage when one channel is anonymized but others are not. Addressing these problems requires bridging the face De-ID literature reviewed in this survey with the soft biometric anonymization literature, an essential direction for future research.

{\section{Conclusion}
	\label{sec:conclusion}}

This survey presents the first unified treatment of face de-identification (De-ID) spanning the full acquisition pipeline, from the physical world through the sensor interface to the digital domain. Our domain-centric analysis exposes a clear trade-off: digital methods offer mature generative fidelity but expose raw identity signals; physical methods push the trust boundary outward yet remain limited by robustness and social acceptability; and sensor-domain methods embed privacy at capture, the closest realization of a privacy-by-design paradigm. Our review of evaluation protocols further reveals substantial fragmentation across datasets, thresholds, and metrics.
Progress will require standardized benchmarks jointly measuring privacy, utility, and quality, together with cross-domain co-design that advances physical materials, programmable optics, and verifiable generative models in tandem. Meeting the emerging challenges of foundation-model-scale recognition and multimodal surveillance will be central to building vision systems that are trustworthy and socially responsible.

\section*{Acknowledgement}
This work was supported by the Research Council of Finland (former Academy of Finland) Academy Professor project EmotionAI (grants 336116, 359894), the University of Oulu \& Research Council of Finland Profi 7 (grant 352788), EU HORIZON-MSCA-SE-2022 project ACMod (grant 101130271), and the Finnish Doctoral Program Network in Artificial Intelligence, AI-DOC (decision number VN/3137/2024-OKM-6). 
As well, the authors wish to acknowledge CSC – IT Center for Science, Finland, for computational resources.

	\small
	\bibliographystyle{IEEEtran}
	\bibliography{reference}

    \vspace{-8mm}
    \begin{IEEEbiography}[{\includegraphics[width=1in,height=1.25in,clip,keepaspectratio]{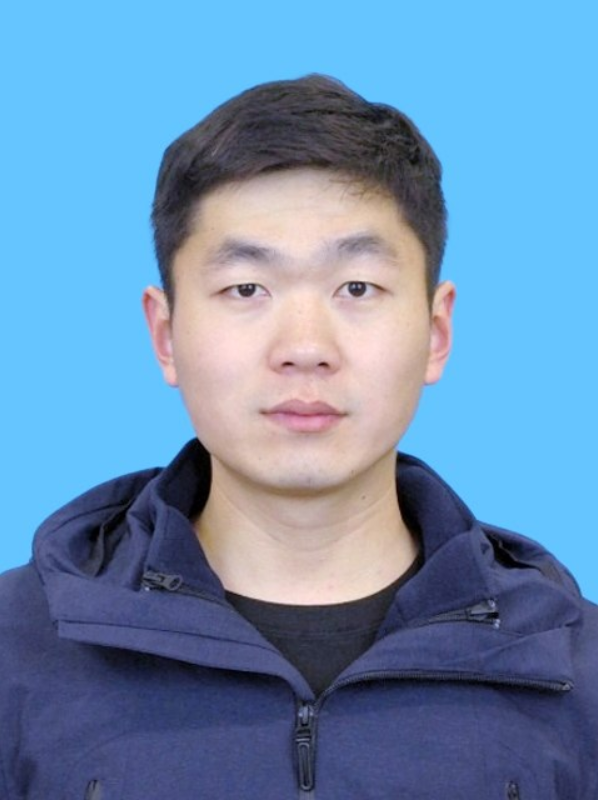}}]{Hui Wei}
		received the PhD degree in computer science from Wuhan University, China. He is currently a postdoctoral researcher with the Center for Machine Vision and Signal Analysis, University of Oulu, Finland. He has authored or coauthored papers in mainstream conferences and journals, including CVPR, NeurIPS, AAAI, ACM Multimedia, and the IEEE TPAMI. His research interests include trustworthy AI, privacy protection, and machine learning. 
	\end{IEEEbiography}
    \vspace{-8mm}

    \begin{IEEEbiography}[{\includegraphics[width=1in,height=1.25in,clip,keepaspectratio]{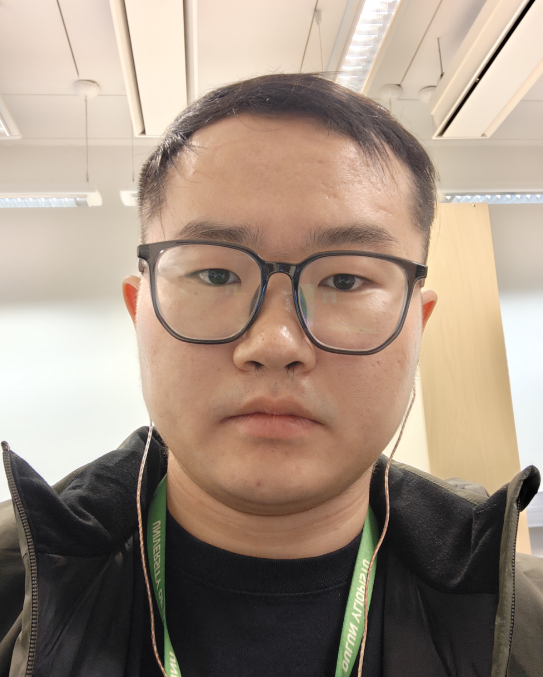}}]{Hao Yu} is currently a PhD student at the Center for Machine Vision and Signal Analysis, University of Oulu, Finland. His research interests focus on visual representation learning, facial attribute analysis, and visual generative models.
    \end{IEEEbiography}
    \vspace{-8mm}

    \begin{IEEEbiography}[{\includegraphics[width=1in,height=1.25in,clip,keepaspectratio]{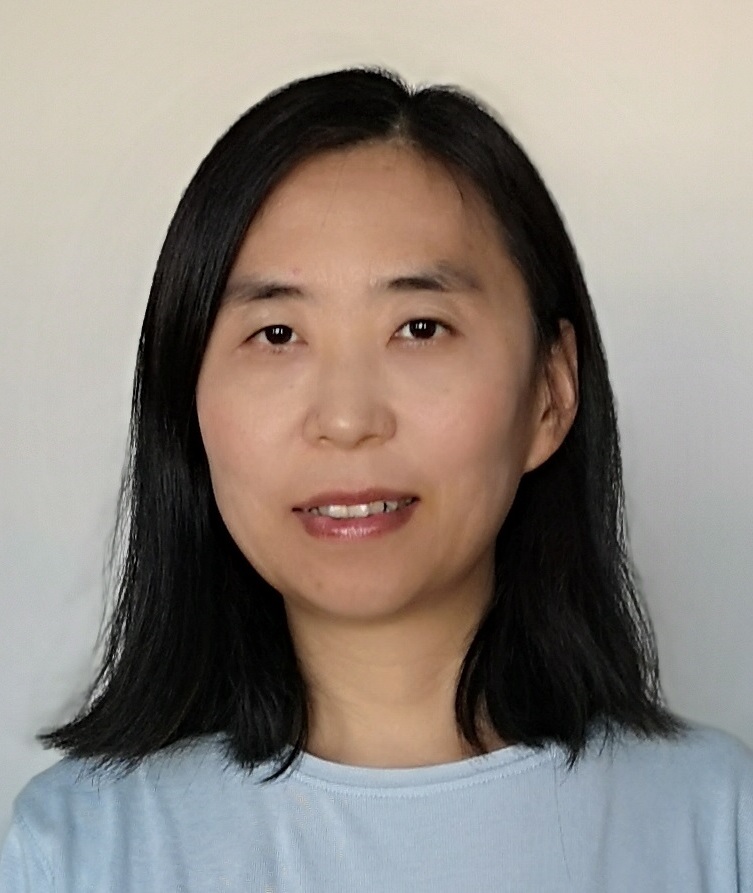}}]{Guoying Zhao} (IEEE Fellow 2022) received the Ph.D. degree in computer science from the Chinese Academy of Sciences, Beijing, China, in 2005. She is currently an Academy Professor and full Professor (tenured in 2017) with University of Oulu, and a PI with ELLIS Institute Finland. She is a member of Academia Europaea, a member of Finnish Academy of Sciences and Letters, Fellow of IEEE, IAPR, ELLIS and AAIA. 
    Her current research interests include image and video representation, facial-expression and micro-expression recognition, emotional gesture analysis, affective computing, and biometrics. 
    \end{IEEEbiography}
	%
	%
	%
	%

	
	
	
\end{document}